\pdfoutput=1
\documentclass[journal]{IEEEtran}
\usepackage[caption=false,font=footnotesize]{subfig}
\usepackage{cite}
\usepackage{amsmath,amssymb,amsfonts}
\usepackage{amsthm}
\usepackage{booktabs,multirow,array,tabularx}
\usepackage{graphicx,adjustbox}
\usepackage{dblfloatfix}
\usepackage{placeins}
\usepackage{pifont}
\usepackage[hidelinks]{hyperref}
\usepackage{xurl}
\usepackage{orcidlink}

\theoremstyle{plain}
\newtheorem{theorem}{Theorem}
\newtheorem{corollary}{Corollary}
\begin{document}
\bstctlcite{BSTcontrol}

\title{Stealthy, High-Impact False Data Injection Attacks in Smart Grid Using Physics-Constrained GAN with Limited System Knowledge}
\author{Md Mainul Islam\orcidlink{0000-0002-9686-5994},~\IEEEmembership{Graduate Student Member, IEEE,}
        Hasan Kurban\orcidlink{0000-0003-3142-2866},
        Abdulrahman Takiddin\orcidlink{0000-0003-4793-003X},~\IEEEmembership{Member, IEEE,}
        Muhammad Ismail\orcidlink{0000-0002-8051-9747},~\IEEEmembership{Senior Member, IEEE,}
        Rachad Atat\orcidlink{0000-0001-8075-6243},~\IEEEmembership{Senior Member, IEEE,}
         Erchin Serpedin\orcidlink{0000-0001-9069-770X},~\IEEEmembership{Fellow,~IEEE}%
\IEEEcompsocitemizethanks{
\IEEEcompsocthanksitem Md. Mainul Islam and Erchin Serpedin are with the Department of Electrical and Computer Engineering, Texas A\&M University, College Station, TX 77843, USA  (e-mails: mdmainul11@tamu.edu; eserpedin@tamu.edu). Hasan Kurban is with the College of Science and Engineering, Hamad Bin Khalifa University, Doha, Qatar (e-mail: hkurban@hbku.edu.qa). Abdulrahman Takiddin is with the Department of Electrical and Computer Engineering, FAMU-FSU College of Engineering, Florida State University, Tallahassee, FL 32310, USA (e-mail: a.takiddin@fsu.edu). Muhammad Ismail is with the Cybersecurity Education, Research, and Outreach Center and the Department of Computer Science, Tennessee Tech University, Cookeville, TN 38505, USA  (e-mail: mismail@tntech.edu).    Rachad Atat is with the Department of Computer Science and Mathematics, Lebanese American University, Beirut 1102 2801, Lebanon  (e-mail: rachad.atat@lau.edu.lb). Md. Mainul Islam is the corresponding author.}
\thanks{This work has been submitted to the IEEE for possible publication. Copyright may be transferred without notice, after which this version may no longer be accessible.}
}

\markboth{}%
{Islam \MakeLowercase{\textit{et al.}}: False Data Injection Attacks in Smart Grid Using DC-Guided Physics-Constrained GAN}
\maketitle

\begin{abstract}
False data injection (FDI) attacks can distort power system state estimation while retaining plausible measurement patterns. Existing FDI methods  trade stealth for impact or vice versa. They often assume access to internal grids, all-bus sensing, or defender models. In this paper, we propose a detector-agnostic physics-constrained generative adversarial network (PCGAN) that learns decoupled power-flow sensitivities from historical measurements of deployed sensors. The learned active-power--angle and reactive-power--voltage-magnitude relationships guide perturbations restricted to compromised channels, while adversarial training promotes realistic spatiotemporal patterns. We implement PCGAN and seven existing FDI attacks on the IEEE 300-bus system using one year of observed hourly demand from PJM Interconnection, a regional transmission organization in the  United States, and synthetic bus-level load profiles. We compare their performance across 10\%, 30\%, 60\%, and 100\% compromised sensors. Attack stealth is assessed against ten machine-learning (ML) detectors and conventional bad-data detection (BDD) methods, while physical impact is quantified through nonlinear alternating-current weighted least-squares state estimation. PCGAN achieves 84.28\% mean ML evasion and 22.31\% mean spatial load misperception across the four scenarios. Its mean stealth--impact trade-off score reaches 18.69\%, compared with 5.61\% for the strongest competing attack. PCGAN also achieves 83.45\% chi-square BDD evasion and 86.02\% largest-normalized-residual BDD evasion. These results demonstrate a favorable stealth--impact balance under varying sensor-access levels and provide challenging attack cases for evaluating FDI detectors.
\end{abstract}


\section{Introduction}

Modern transmission grids rely on state estimation driven by supervisory control and data acquisition (SCADA) and phasor measurement unit (PMU) measurements to support dispatch, contingency analysis, congestion monitoring, and operator decision-making. The integrity of these measurements is therefore directly linked to the reliability of downstream control and planning functions. In a false data injection (FDI) attack, an adversary modifies a subset of the telemetry before it is processed by the control center. Conventional bad-data detection (BDD) identifies gross errors, but coordinated perturbations that remain close to normal operating relationships are substantially harder to detect, particularly as learning-based detectors become more widespread.

A meaningful FDI attack must balance two conflicting objectives. Small perturbations may remain undetected but produce little physical consequence, whereas large perturbations cause greater impact but become conspicuous. Detector evasion rate alone therefore does not fully characterize attack strength, and a credible attack should achieve a favorable \emph{stealth--impact trade-off}. In addition, model-based attacks often assume detailed knowledge of the grid and estimator, and adversarial-learning attacks often assume access to the detector or its gradients. Such assumptions can overstate the information available to an attacker. A further impractical assumption is that sensors are deployed at every bus and that the attacker can write to all of them. Budget-constrained operators instead instrument only selected critical buses, and an external attacker rarely gains write access to every deployed channel. Overlooking this restricted-access setting can overstate attack reach, because perturbing only compromised measurements must remain consistent with the unaltered readings that the state estimator still receives. This study therefore investigates whether an attacker without internal grid or defender-model information, and with write access to only a subset of deployed sensors, can exploit a learned power-flow approximation to generate statistically plausible perturbations with measurable effects on alternating-current (AC) state estimation.

\subsection{Related Work}
\label{sec:related_work}

FDI attacks have been studied from both defensive and offensive perspectives. FDI detection has been studied using model-based, statistical, and learning-based methods, including graph-based models that exploit the network structure, whereas attack-generation studies design perturbations that evade such detectors and corrupt state estimation.

\textbf{Model-Based and Data-Driven Detection:} Model-based approaches exploit state-estimation residuals, observability, sparsity, interval models, or other physical consistency relations \cite{huang2021false,luo2021interval,xie2021novel}. Data-driven approaches learn normal and malicious patterns from measurements using semi-supervised learning, recurrent models, convolutional models, federated learning, and distributed detection \cite{zhang2021detecting,yang2021deep,wang2022kfrnn,zhang2022spatio,yin2022subgrid,li2022secure,li2024end,yin2022powerfdnet}. Generative models have also been used defensively for FDI detection \cite{xie2024bayesian,feng2024detection}, and generative adversarial networks (GANs) can synthesize realistic and physically meaningful power-system measurements \cite{zheng2021generative}. However, the robustness of these detectors depends on the attack distributions used in training and evaluation.

\textbf{Graph-Based Detection:} Because power-system measurements are coupled through the network structure, graph neural networks (GNNs) have been used for topology-aware detection, joint detection and localization, generalization to unseen attacks and topologies, and attention- or centrality-based aggregation \cite{boyaci2022graph,boyaci2022joint,li2023graph,chen2023detection,takiddin2023generalized,peng2023localizing,su2024damgat,elnour2025eigenvector,atat2025graphon,aboelmagd2026simplicial}. Robust graph autoencoders address unseen attacks and poisoned training data \cite{takiddin2024robust}, and physics-informed graph transformers use  AC consistency features \cite{elnour2026joint}. These advances strengthen the defender but do not establish robustness against an attacker that adapts its perturbations to remain stealthy and consequential.

\noindent\textbf{Attack Generation:}
Several studies investigated model-based and data-driven FDI attacks under varying grid-knowledge assumptions \cite{jin2019power,lakshminarayana2021data,du2021targeted}. Aboelmagd et al. \cite{aboelmagd2026simplicial} showed that topology-aware attacked-bus selection substantially degrades detection performance, assuming the attacker knows the grid topology. They compared bus-selection strategies under fixed perturbations but did not compare perturbation methods on the same compromised buses or optimize the perturbations.

Learning-based approaches include regional AC attacks \cite{jiao2021new}, Wasserstein GAN-based estimator approximation \cite{costilla2023attack}, surrogate-model-based attacks \cite{rahman2024adversarial}, and a long short-term memory (LSTM) autoencoder GAN (LSTMAE-GAN), which exploits temporal dependencies and estimator-mimicking regularization to generate sparse AC attacks \cite{narang2024physical}. Takiddin \emph{et al.} studied adversarial evasion of learning-based detectors using spatiotemporal features \cite{takiddin2024spatiotemporal}, while Tian \emph{et al.} proposed LESSON, which uses Jacobian and detector knowledge to evade BDD and neural attack localization \cite{tian2024lesson}. 

Tensor-completion-based attacks exploit historical patterns but require grid parameters \cite{liu2024tensor}, whereas black-box GAN attacks on distributed power-supply measurements target data-driven control \cite{liu2025false}. More recently, representation-learning GAN \cite{liu2026rlgan} and GAN-based variants \cite{nawaz2025gan} generated complete measurement vectors from historical data without explicitly enforcing AC power-flow constraints. However, their effectiveness under restricted sensor access remains uncertain, as modifying only compromised measurements may disrupt learned dependencies and reduce stealth.

\textbf{Limitations of Existing Work:}
The limitations of existing FDI attack-generation approaches are summarized as follows:
\begin{enumerate}
    \item \emph{Sensor deployment and attacker access:} The works \cite{jin2019power,lakshminarayana2021data,du2021targeted,jiao2021new,costilla2023attack,rahman2024adversarial,narang2024physical,takiddin2024spatiotemporal,tian2024lesson,liu2025false,liu2026rlgan,nawaz2025gan} assume every bus is instrumented, and \cite{lakshminarayana2021data,costilla2023attack,rahman2024adversarial,tian2024lesson,liu2026rlgan,nawaz2025gan} perturb measurements at all buses, which is impractical for budget-constrained grids that strategically deploy sensors only at selected critical locations, providing full or partial observability.

    \item \emph{Grid knowledge:} The works \cite{jin2019power,tian2024lesson,liu2024tensor} require the Jacobian, topology, or line parameters; \cite{du2021targeted,jiao2021new,narang2024physical} require the connectivity of the attack region; and the strongest attacker in \cite{takiddin2024spatiotemporal, aboelmagd2026simplicial} is an insider who knows the full topology. Such internal grid information is rarely available to an external attacker.

    \item \emph{Defender dependence:} The works \cite{tian2024lesson,takiddin2024spatiotemporal,liu2025false} require the parameters, gradients, or input--output responses of the targeted detector or model, which couples stealth to that model; transferability to unseen detectors is examined only partially \cite{takiddin2024spatiotemporal,liu2025false}.

    \item \emph{Stealth--impact trade-off:} Most studies \cite{jin2019power,lakshminarayana2021data,du2021targeted,jiao2021new,costilla2023attack,rahman2024adversarial,narang2024physical,liu2024tensor,takiddin2024spatiotemporal,tian2024lesson,liu2025false,liu2026rlgan} report stealth, evasion success, or estimation and prediction errors rather than physical or economic consequences. Only \cite{nawaz2025gan} reports energy theft for a single 5-bus case. The trade-off between stealth and grid impact of FDI attacks therefore remains unexplored.
\end{enumerate}

Overall, the joint relationship among  physics consistency, conventional and heterogeneous learning-based detector evasion, and grid impact under limited attacker knowledge and restricted sensor access remains insufficiently characterized.

\subsection{Our Contributions}
\label{sec:contributions}

To address the identified gap, this paper proposes a direct-current (DC)-guided physics-constrained GAN (PCGAN) that identifies approximate power-flow relationships from historical measurements of restricted sensors  without accessing the network admittance matrix, the defender's estimator, or its detectors. Following the fast-decoupled interpretation, the learned model relates active power to voltage angles and reactive power to voltage magnitudes. These relationships guide perturbations on writable measurements, while adversarial training encourages statistical and temporal resemblance to benign operation. The resulting construction is constrained by the learned relationships rather than the complete nonlinear AC equations. The main contributions are as follows.
\begin{itemize}
\item A detector-agnostic generative attack that learns approximate active-power--angle and reactive-power--voltage-magnitude relationships from the historical measurements of a subset of sensors, without requiring the network admittance matrix or access to the deployed detectors.
\item A physics-guided perturbation construction that coordinates voltage-state changes and compromised sensor-owned measurements while using adversarial distribution learning to preserve the statistical and temporal characteristics of benign operation.
\item A mathematical analysis of the learned model and its interaction with nonlinear AC state estimation, including restricted measurement consistency, conditional residual bounds, and a comparison with equal-energy random perturbations.
\item An evaluation across eight attack families, ten heterogeneous detectors, and nonlinear AC weighted least-squares (WLS) estimation establishes the effectiveness of PCGAN in jointly maintaining stealth and inducing spatial-load misperception. Unlike competing attacks that achieve either high evasion or substantial estimation bias, PCGAN provides a more favorable balance between these objectives, highlighting the limitations of existing  detectors against physics-informed  adversarial perturbations.
\end{itemize}

The remainder of the paper is organized as follows. Section~II presents the system and threat model, Section~III describes PCGAN, Sections~IV and V present the benchmarks and experimental methodology, Section~VI presents the results and discussion, and Section~VII concludes the paper.

\section{System and Threat Model}
\label{sec:system_threat}

\subsection{AC Power System and State Estimation}
\label{sec:se}

Consider an AC transmission network with bus set $\mathcal B$. Bold symbols denote vectors and matrices and $\top$ denotes transposition. The complex voltage at bus $i\in\mathcal B$ is $v_i=V_i e^{\jmath\theta_i}$, where $V_i$ is the magnitude, $\theta_i$ the angle, and $\jmath=\sqrt{-1}$; the vectors $\mathbf v$, $\mathbf V$, and $\boldsymbol\theta$ collect these quantities in bus order. The $(i,k)$ entry of the bus-admittance matrix $\mathbf Y$ is $Y_{i,k}=G_{i,k}+\jmath B_{i,k}$, with conductance $G_{i,k}$ and susceptance $B_{i,k}$, and $\theta_{i,k}=\theta_i-\theta_k$. The AC nodal power-balance equations are
\begin{align}
\label{eq:pf}
P_i&=V_i\sum_{k\in\mathcal B}V_k\big(G_{i,k}\cos\theta_{i,k}+B_{i,k}\sin\theta_{i,k}\big),\\
Q_i&=V_i\sum_{k\in\mathcal B}V_k\big(G_{i,k}\sin\theta_{i,k}-B_{i,k}\cos\theta_{i,k}\big),
\end{align}
where $P_i=P_i^{\mathrm g}-P_i^{\mathrm d}$ and $Q_i=Q_i^{\mathrm g}-Q_i^{\mathrm d}$ are the net active and reactive injections, with $P_i^{\mathrm g},Q_i^{\mathrm g}$ the generation and $P_i^{\mathrm d},Q_i^{\mathrm d}$ the demand at bus $i$. Collecting the injections in $\mathbf P$ and $\mathbf Q$, these read $\mathbf P+\jmath\mathbf Q=\mathbf v\odot(\mathbf Y\mathbf v)^{*}$, where $\odot$ is the element-wise product and $(\cdot)^{*}$ the complex conjugate. The grid operates over hourly intervals indexed by $t$ with time-varying loading; the measurements collected over these intervals form the historical data available for learning.

Let $\mathbf m_t\in\mathbb R^{m}$ denote the $m$-channel measurement vector containing voltage magnitudes, bus injections, and branch flows available to the control center at time $t$. The measurement model is $\mathbf m_t=\mathbf h(\mathbf x_t)+\boldsymbol\nu_t$, where $\mathbf h:\mathbb R^{n_x}\to\mathbb R^m$ is the AC measurement function, $n_x$ is the number of state variables, and $\boldsymbol\nu_t$ is zero-mean noise with covariance $\mathbf R=\operatorname{diag}(\sigma_1^2,\ldots,\sigma_m^2)$. Here, $\sigma_j>0$ denotes the standard deviation (SD) of measurement channel $j$. Let $\bar{\boldsymbol\theta}_t$ denote the voltage-angle vector excluding the fixed reference-bus angle. The state vector is then $\mathbf x_t=[\bar{\boldsymbol\theta}_t^{\top},\mathbf V_t^{\top}]^{\top}$. The control center solves the AC-WLS problem
\begin{equation}
\label{eq:wls}
\widehat{\mathbf x}_t=\arg\min_{\mathbf x}J_w(\mathbf m_t,\mathbf x),~
J_w(\mathbf m,\mathbf x)=\|\mathbf R^{-1/2}[\mathbf m-\mathbf h(\mathbf x)]\|_2^2,
\end{equation}
where $\widehat{\mathbf x}_t$ is the estimate and $\|\cdot\|_2$ denotes the Euclidean norm. The problem is solved iteratively by Gauss--Newton updates using the measurement Jacobian $\mathbf H(\mathbf x)=\partial\mathbf h(\mathbf x)/\partial\mathbf x\in\mathbb R^{m\times n_x}$. A BDD test flags the sample when $J_w(\mathbf m_t,\widehat{\mathbf x}_t)$ exceeds a chi-square threshold $\tau_0$. An attack that shifts the estimate while passing this test is stealthy under BDD.

\subsection{Threat Model}
\label{sec:threat}

\begin{figure}[!t]
  \centering
  \subfloat[]{\includegraphics[width=\linewidth]{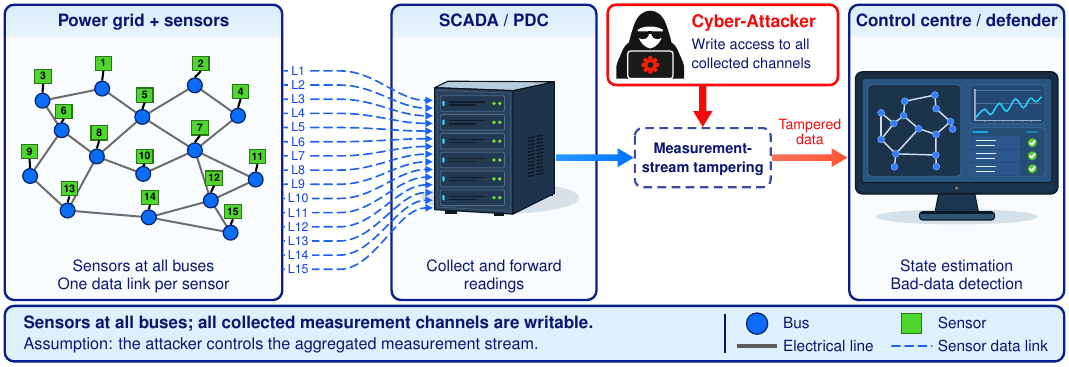}%
    \label{fig:fdi_full}}\\
  \subfloat[]{\includegraphics[width=\linewidth]{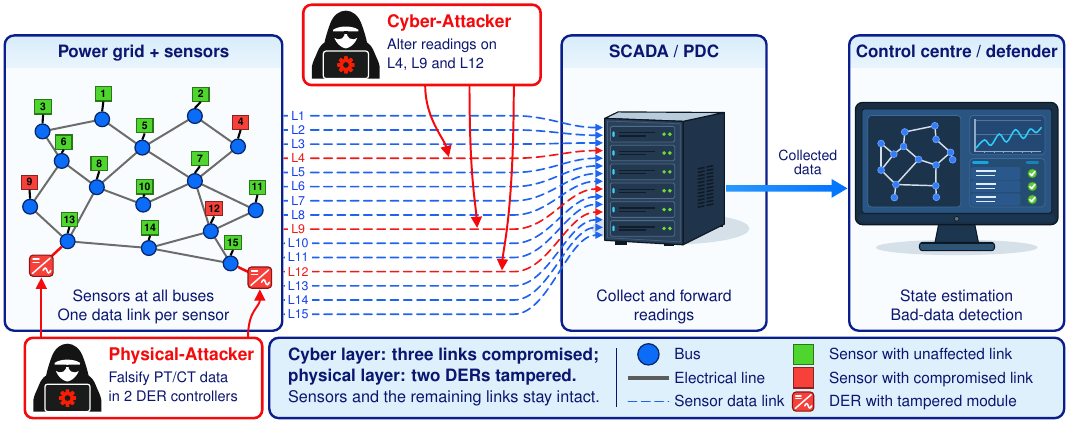}%
    \label{fig:fdi_subset}}\\
  \subfloat[]{\includegraphics[width=\linewidth]{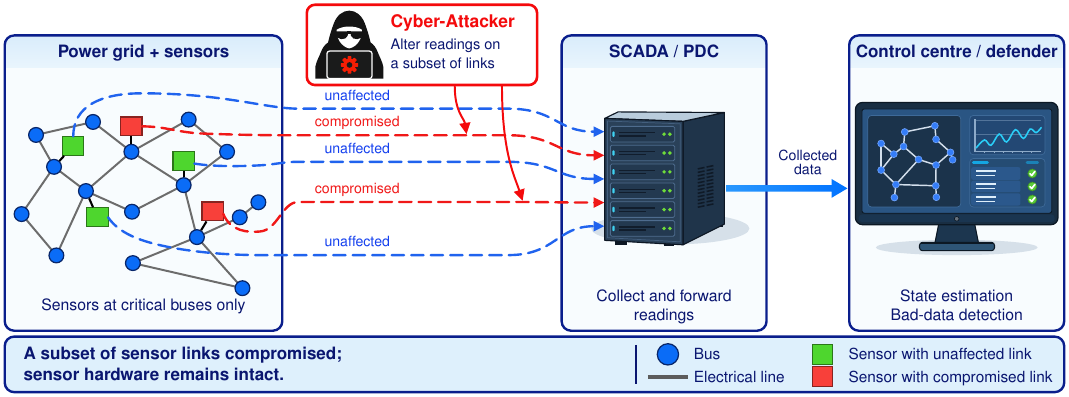}%
    \label{fig:fdi_sparse}}
  \caption{FDI attack scenarios of (a) \cite{lakshminarayana2021data,costilla2023attack,rahman2024adversarial,tian2024lesson,liu2026rlgan,nawaz2025gan}, (b) \cite{jin2019power,du2021targeted,jiao2021new,narang2024physical,takiddin2024spatiotemporal,liu2025false,xie2024bayesian,feng2024detection,takiddin2023generalized,takiddin2024robust,elnour2025eigenvector,aboelmagd2026simplicial}, and (c) this work and \cite{liu2024tensor,elnour2026joint}.}
  \label{fig:fdi_scenarios}
\end{figure}

Let $\mathcal S\subseteq\mathcal B$ be the set of sensor buses, $n_s=|\mathcal S|$, and $\mathcal A\subseteq\mathcal S$ the compromised sensor buses. Fig.~\ref{fig:fdi_scenarios} compares three FDI threat models. In (a), every bus is measured ($n_s=|\mathcal B|$) and the attacker alters all measurements ($\mathcal A=\mathcal S$); the learning-based attacks of this type \cite{costilla2023attack,rahman2024adversarial,liu2026rlgan,nawaz2025gan} produce complete measurement vectors over all buses, injected as a whole or, in \cite{rahman2024adversarial}, after removing entries flagged by a substitute detector. When $n_s<|\mathcal B|$, such a vector does not match the control center's measurement configuration, and restricting it to existing channels still requires write access to every sensor. In (b), every bus is measured but the attacker alters a subset ($\mathcal A\subset\mathcal S$), e.g., through selected sensor links or tampered modules of distributed energy resources \cite{liu2025false}. These attacks are still constructed with measurements at every bus of the network or attack region, which may not exist when $n_s<|\mathcal B|$. Several such works complete the all-bus vector with pseudo-measurements \cite{nawaz2025gan,elnour2025eigenvector}, which are generated inside the control center and cannot be altered through communication links. In (c), only $n_s<|\mathcal B|$ buses host sensors and the attacker compromises $\mathcal A\subset\mathcal S$. This is the most practical setting, since budget-constrained operators instrument only critical buses and an external attacker rarely controls every channel. Therefore, this study adopts Fig.~\ref{fig:fdi_sparse}.

Each sensor bus $i\in\mathcal S$ hosts a remote terminal unit (RTU) and a PMU. The RTU reports $V_i$, $P_i$, $Q_i$, and the flows at incident branch ends through SCADA; the PMU reports the voltage phasor $V_i\angle\theta_i$ and the incident current phasors $I_{ij}\angle\phi_{ij}$, with magnitude $I_{ij}$ and angle $\phi_{ij}$ on branch $(i,j)$, through a phasor data concentrator, from which the branch-end flows $P_{ij}=V_iI_{ij}\cos(\theta_i-\phi_{ij})$ and $Q_{ij}=V_iI_{ij}\sin(\theta_i-\phi_{ij})$ follow. Together they form $\mathbf m_t$ of magnitudes, injections, and branch-end flows at the sensor buses. The attacker observes present and historical readings of the compromised sensors only, namely $V_i$, $\theta_i$, $P_i$, $Q_i$, $I_{ij}$, and the owned branch-end flows for $i\in\mathcal A$, collected as $\mathbf z_t\in\mathbb R^n$, whose $n$ channels need not match those of $\mathbf m_t$. The attacker has neither the network model (topology, line parameters, $\mathbf Y$, or $\mathbf H$) nor the estimator and detector parameters; PCGAN instead identifies approximate power-flow sensitivities from benign observations.

Each $i\in\mathcal A$ gives the attacker write access to the channels of $\mathbf m_t$ its RTU and PMU report, but not to those of uncompromised sensors. The attacked measurement is $\mathbf m_t^{\mathrm a}=\mathbf m_t+\mathbf a_t$, where the attack $\mathbf a_t\in\mathbb R^m$ is zero outside the writable channels; protected flows related to an altered injection can therefore reveal inconsistencies. The objective is to shift the estimate from its benign value $\widehat{\mathbf x}_t^{\mathrm b}$ to an attacked value $\widehat{\mathbf x}_t^{\mathrm a}$, distorting the control center's view of the load distribution, while evading the evaluated detectors and the AC-WLS residual test.

\section{Proposed Physics-Constrained GAN}
\label{sec:pcgan}

\begin{figure*}[!t]
\centering
\includegraphics[width=0.74\textwidth]{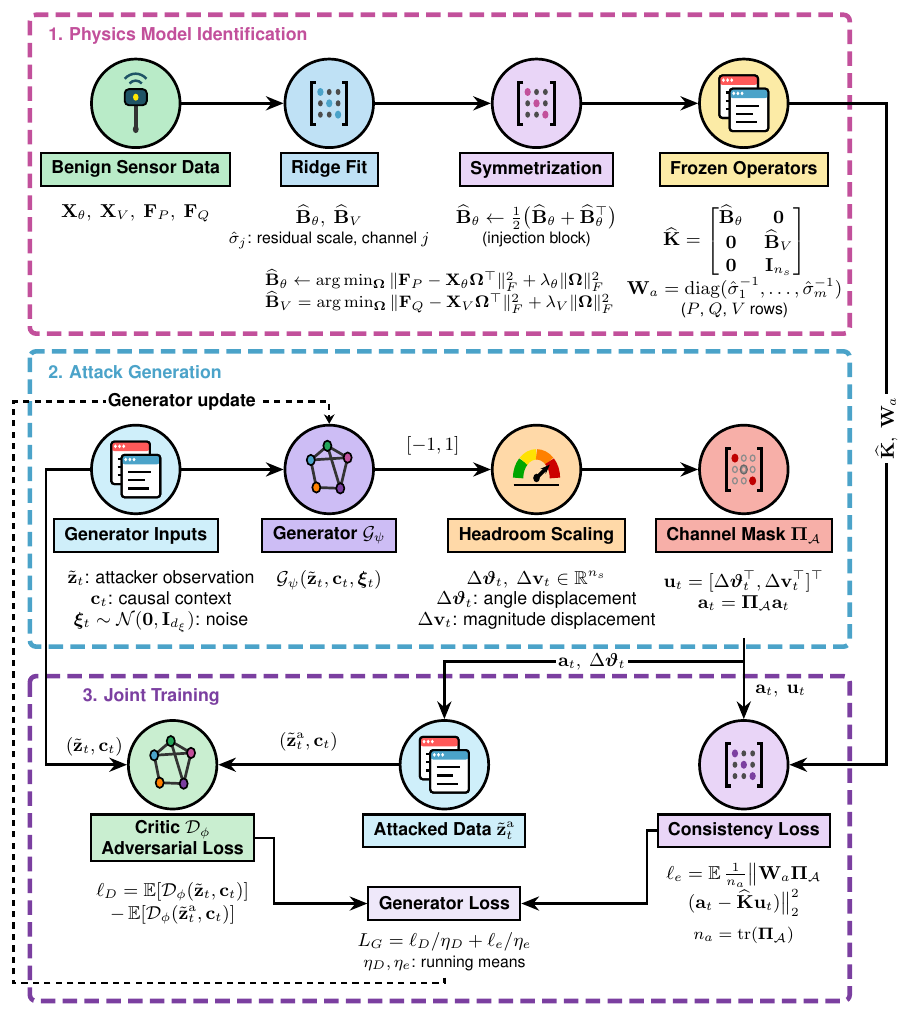}
\caption{Architecture of the proposed PCGAN.}
\label{fig:pcgan}
\end{figure*}

PCGAN first identifies a decoupled measurement model from benign sensor data and then uses it to guide attack generation. A consistency penalty encourages agreement with the learned sensitivities, while adversarial learning shapes the perturbations written to compromised channels. \emph{Physics-constrained} refers to this empirical, model-guided construction; it does not imply satisfaction of the nonlinear AC equations or guaranteed evasion of AC-WLS BDD. Fig.~\ref{fig:pcgan} illustrates the three stages of PCGAN, namely physics model identification, attack generation, and joint training of the generator and critic, together with the data flow among them.

\subsection{Physics Model Identification}
\label{sec:physics_id}

Without topology, line parameters, or $\mathbf Y$, the attacker cannot form the derivatives of~\eqref{eq:pf}, and estimating unrestricted AC relationships from partial observations is ill-posed because active and reactive powers depend jointly on angles and magnitudes with operating-point-dependent sensitivities. PCGAN therefore identifies a decoupled linear approximation from its observed benign measurements, including branch-end flows; the control center retains the nonlinear estimator in~\eqref{eq:wls}. For full-network changes $\Delta\mathbf P$, $\Delta\mathbf Q$, $\Delta\bar{\boldsymbol\theta}$, and $\Delta\mathbf V$ of injections, reduced angles, and magnitudes about an operating point, a first-order expansion gives
\begin{equation}
\label{eq:ac_linearization}
\begin{bmatrix}\Delta\mathbf P\\\Delta\mathbf Q\end{bmatrix}
\simeq
\begin{bmatrix}\mathbf J_1&\mathbf J_2\\\mathbf J_3&\mathbf J_4\end{bmatrix}
\begin{bmatrix}\Delta\bar{\boldsymbol\theta}\\\Delta\mathbf V\end{bmatrix},
\end{equation}
with Jacobian blocks $\mathbf J_1=\partial\mathbf P/\partial\bar{\boldsymbol\theta}$, $\mathbf J_2=\partial\mathbf P/\partial\mathbf V$, $\mathbf J_3=\partial\mathbf Q/\partial\bar{\boldsymbol\theta}$, and $\mathbf J_4=\partial\mathbf Q/\partial\mathbf V$. Under fast-decoupled assumptions (small angle differences, weak resistance, voltages near nominal) the cross blocks are neglected, $\Delta\mathbf P\simeq\mathbf J_1\Delta\bar{\boldsymbol\theta}$ and $\Delta\mathbf Q\simeq\mathbf J_4\Delta\mathbf V$; only the first relation corresponds to lossless DC power flow. \emph{DC-guided} thus denotes a learned, approximately decoupled P--angle/Q--magnitude construction, not an exact network model.

The attacker learns responses of \emph{monitored power channels} rather than all nodal injections. Let $\mathbf h_P$ and $\mathbf h_Q$ select the monitored active- and reactive-power measurements (injections and branch-end flows), with derivatives $\mathbf H_P=\partial\mathbf h_P/\partial\bar{\boldsymbol\theta}$ and $\mathbf H_Q=\partial\mathbf h_Q/\partial\mathbf V$ at the operating point, and let $\mathbf C_\theta\in\mathbb R^{(|\mathcal B|-1)\times n_s}$ and $\mathbf C_V\in\mathbb R^{|\mathcal B|\times n_s}$ embed sensor-coordinate changes into the network state with unobserved changes set to zero. The physical sensitivities restricted to observed coordinates are $\widetilde{\mathbf B}_\theta=\mathbf H_P\mathbf C_\theta$ and $\widetilde{\mathbf B}_V=\mathbf H_Q\mathbf C_V$; they are neither the full Jacobian nor the susceptance matrix and are used only in the analysis.

PCGAN estimates its sensitivities from $N$ benign training observations. With $n_P$ and $n_Q$ the numbers of monitored active- and reactive-power channels, $\mathbf X_\theta,\mathbf X_V\in\mathbb R^{N\times n_s}$ contain the centered observed sensor angles and magnitudes (zero for unobserved sensors), and $\mathbf F_P\in\mathbb R^{N\times n_P}$, $\mathbf F_Q\in\mathbb R^{N\times n_Q}$ the centered monitored power measurements. The operators are fitted once by ridge regression over the coefficient matrix $\boldsymbol\Omega$ and then frozen:
\begin{align}
\label{eq:fit_dc}
\widehat{\mathbf B}_\theta&\leftarrow
\arg\min_{\boldsymbol\Omega}
\|\mathbf F_P-\mathbf X_\theta\boldsymbol\Omega^\top\|_F^2+\lambda_\theta\|\boldsymbol\Omega\|_F^2,\nonumber\\
\widehat{\mathbf B}_V&=
\arg\min_{\boldsymbol\Omega}
\|\mathbf F_Q-\mathbf X_V\boldsymbol\Omega^\top\|_F^2+\lambda_V\|\boldsymbol\Omega\|_F^2,
\end{align}
where $\|\cdot\|_F$ is the Frobenius norm and $\lambda_\theta,\lambda_V>0$ are regularization parameters. After fitting, the square active-injection block of $\widehat{\mathbf B}_\theta$ is symmetrized by averaging it with its transpose, as a DC susceptance matrix should be; the flow rows and $\widehat{\mathbf B}_V$ are unchanged, and $\widehat{\mathbf B}_\theta$ denotes the symmetrized operator hereafter. The per-channel scales $\hat\sigma_j$ of the fitting residuals are retained as an empirical diagonal whitening for the consistency penalty below.

The learned sensitivities need not equal their restricted physical counterparts because unobserved states vary with measured states. Let $\mathbf T_\theta\in\mathbb R^{(|\mathcal B|-1)\times n_s}$ and $\mathbf T_V\in\mathbb R^{|\mathcal B|\times n_s}$ be the least-squares maps from observed sensor-coordinate changes to full-network angle and magnitude changes in the benign data, and let $\mathbf T$ stack them so that an attacker-side coordinate change maps to the full state change $\Delta\mathbf x=\mathbf T\mathbf u$. In a noiseless, exactly decoupled, fixed-Jacobian regime with full-column-rank designs and vanishing regularization, the unsymmetrized coefficients equal $\mathbf H_P\mathbf T_\theta$ and $\mathbf H_Q\mathbf T_V$; symmetrization, noise, regularization, and model mismatch move the learned operators away from these values. These maps tie the identified model to the attack construction of Section~\ref{sec:attack_gen}. The displacement $\mathbf u_t$ generated there induces the state change $\Delta\mathbf x_t=\mathbf T\mathbf u_t$, and the perturbation guided by $\widehat{\mathbf K}\mathbf u_t$ approximates the measurement change $\mathbf H\Delta\mathbf x_t$ that this state change would produce, which is exactly a perturbation that the linearized estimator can absorb. The gap $\widehat{\mathbf K}-\mathbf H\mathbf T$ therefore measures how far the guided perturbation departs from such an absorbable one, and Section~\ref{sec:properties} uses it to bound the residual visibility of PCGAN. The maps $\mathbf T_\theta$ and $\mathbf T_V$ are unavailable to the attacker and appear only in the analysis.

\subsection{Attack Generation and Objective}
\label{sec:attack_gen}

The generator $\mathcal G_\psi$ and the Wasserstein critic $\mathcal D_\phi$ have parameters $\psi$ and $\phi$. The generator receives the standardized attacker-side observation $\tilde{\mathbf z}_t$, a causal context $\mathbf c_t$ of lagged benign observations, and a noise vector $\boldsymbol\xi_t\sim\mathcal N(\mathbf0,\mathbf I_{d_\xi})$, where $\mathcal N$ denotes the normal distribution, $d_\xi$ the noise dimension, and $\mathbf I_{d_\xi}$ the identity; it returns bounded directions in $[-1,1]$ for the angle and magnitude coordinates of the compromised sensors and for the writable measurement channels. Each direction component is scaled by the signed headroom of its coordinate to the historical calibration envelope, keeping attacked readings within the benign range. This yields angle and magnitude displacements $\Delta\boldsymbol\vartheta_t,\Delta\mathbf v_t\in\mathbb R^{n_s}$, zero-padded at uncompromised sensors and stacked as $\mathbf u_t=[\Delta\boldsymbol\vartheta_t^\top,\Delta\mathbf v_t^\top]^\top$, which is an attacker-side coordinate change rather than a full AC state change, and the written perturbation $\mathbf a_t$. With $\boldsymbol\Pi_{\mathcal A}\in\mathbb R^{m\times m}$ the diagonal mask of the magnitude, injection, and branch flow channels of compromised sensors, $\mathbf a_t=\boldsymbol\Pi_{\mathcal A}\mathbf a_t$.

Let $\widehat{\mathbf K}\in\mathbb R^{m\times 2n_s}$ map $\mathbf u_t$ to the SCADA measurement ordering. Its active- and reactive-power rows contain the unit-converted $\widehat{\mathbf B}_\theta$ and $\widehat{\mathbf B}_V$, and its voltage-magnitude rows select $\Delta\mathbf v_t$, so that, up to the ordering of the measurement channels,
\begin{equation}
\label{eq:khat}
\widehat{\mathbf K}=\begin{bmatrix}\widehat{\mathbf B}_\theta&\mathbf 0\\ \mathbf 0&\widehat{\mathbf B}_V\\ \mathbf 0&\mathbf I_{n_s}\end{bmatrix}.
\end{equation} The decoupled model is not imposed as a hard product; instead the generator is trained to make its written perturbation consistent with it through the whitened consistency loss
\begin{equation}
\label{eq:pcgan_phy}
\ell_e=\mathbb E\,\tfrac{1}{n_a}\big\|\mathbf W_a\boldsymbol\Pi_{\mathcal A}(\mathbf a_t-\widehat{\mathbf K}\mathbf u_t)\big\|_2^2,
\end{equation}
where $\mathbb E$ denotes expectation over training samples, $n_a=\operatorname{tr}(\boldsymbol\Pi_{\mathcal A})$ is the number of writable channels, and $\mathbf W_a=\operatorname{diag}(\hat\sigma_1^{-1},\ldots,\hat\sigma_m^{-1})$ is the diagonal whitening from the identification residuals. No condition is enforced at protected injection or flow channels, which can therefore expose inconsistencies through the defender's nonlinear residual.

The standardized attacker-side observation after applying $\mathbf a_t$, shifting the compromised angles by $\Delta\boldsymbol\vartheta_t$, and recomputing derived current magnitudes is $\tilde{\mathbf z}_t^{\mathrm a}$. With the same context for benign and attacked samples, the critic score difference is
\begin{equation}
\label{eq:pcgan_adv}
\ell_D=\mathbb E[\mathcal D_\phi(\tilde{\mathbf z}_t,\mathbf c_t)]-\mathbb E[\mathcal D_\phi(\tilde{\mathbf z}_t^{\mathrm a},\mathbf c_t)],
\end{equation}
and the critic is trained with a gradient penalty. 

The generator minimizes an aggregate loss function
\begin{equation}
\label{eq:pcgan_obj}
L_G=\ell_D/\eta_D+\ell_e/\eta_e,
\end{equation}
where $\eta_D$ and $\eta_e$ are the running means of $|\ell_D|$ and $|\ell_e|$ over all previous generator updates, treated as constants in the gradient, so that both terms enter with comparable magnitude. The consistency loss guides perturbations toward the identified model, and the critic favors benign-like measurement histories.

The critic and generator alternate updates; an exponential moving average of generator parameters is kept, and validation critic gaps select the checkpoint. The identified operators, calibration envelopes, and normalization statistics are frozen for held-out testing.

\subsection{Mathematical Properties}
\label{sec:properties}
\label{sec:pcgan_properties}

The following results connect the learned perturbations to WLS detection. We first characterize linearized residual visibility, then compare equal-energy attacks and bound the nonlinear residual.

Let $\widehat{\mathbf x}$ be an interior stationary point of benign AC-WLS. Define $\mathbf W=\mathbf R^{-1/2}$, $\mathbf H=\mathbf H(\widehat{\mathbf x})$, $\mathbf r=\mathbf W[\mathbf m-\mathbf h(\widehat{\mathbf x})]$, and $J_0=\|\mathbf r\|_2^2$. Let $\mathbf U=(\mathbf W\mathbf H)(\mathbf W\mathbf H)^\dagger$ be the orthogonal projector onto the column space of $\mathbf W\mathbf H$, where $\dagger$ denotes the Moore--Penrose pseudoinverse. Set $\mathbf P_\perp=\mathbf I_m-\mathbf U$, with $\mathbf I_m$ the $m\times m$ identity.

\begin{theorem}[Residual visibility of FDI attacks]
\label{thm:increment}
For any attack $\mathbf a$, the reoptimized linearized WLS objective $J_\ell(\mathbf a)=\min_{\mathbf d}\|\mathbf r+\mathbf W\mathbf a-\mathbf W\mathbf H\mathbf d\|_2^2$, with $\mathbf d$ denoting a state adjustment, satisfies
\begin{equation}
\label{eq:increment}
J_\ell(\mathbf a)-J_0
=\|\mathbf P_\perp\mathbf W\mathbf a\|_2^2
+2\mathbf r^\top\mathbf P_\perp\mathbf W\mathbf a.
\end{equation}
Hence, $\mathbf P_\perp\mathbf W\mathbf a=\mathbf0$ guarantees unchanged linearized WLS residuals. If $\mathbf W$ and $\mathbf H$ are fixed, $\mathbb E[\mathbf r]=\mathbf0$, and $\mathbf r$ is independent of $\mathbf a$, the expected residual increment equals $\mathbb E\|\mathbf P_\perp\mathbf W\mathbf a\|_2^2$.
\end{theorem}
\begin{IEEEproof}
WLS stationarity implies $\mathbf U\mathbf r=\mathbf0$. Thus, $J_\ell(\mathbf a)=\|\mathbf P_\perp(\mathbf r+\mathbf W\mathbf a)\|_2^2$. Expanding gives~\eqref{eq:increment}; the cross term vanishes in expectation under the stated assumptions.
\end{IEEEproof}

Only the component of an attack that cannot be explained by a linearized state adjustment contributes additional residual energy. Under a fixed affine Gaussian model with deterministic attacks, this energy determines the noncentrality of the chi-square BDD statistic. Lower projected energy favors BDD evasion under those assumptions.

For PCGAN, write $\mathbf a=\boldsymbol\Pi_{\mathcal A}\widehat{\mathbf K}\mathbf u+\mathbf e$, where $\boldsymbol\Pi_{\mathcal A}$ masks writable channels, $\widehat{\mathbf K}$ is the learned sensitivity, $\mathbf u$ is the generated displacement, and $\mathbf e=\boldsymbol\Pi_{\mathcal A}\mathbf e$ is the consistency error. With $\mathbf T$ the sensor-to-state map of Section~\ref{sec:physics_id}, define $\mathbf E=\widehat{\mathbf K}-\mathbf H\mathbf T$. The attacker does not know $\mathbf H$ or $\mathbf T$.

\begin{theorem}[Stealth advantage over random attacks]
\label{thm:random}
Define three error components, $\mathbf q_1=\boldsymbol\Pi_{\mathcal A}\mathbf E\mathbf u$ (model mismatch), $\mathbf q_2=-(\mathbf I_m-\boldsymbol\Pi_{\mathcal A})\mathbf H\mathbf T\mathbf u$ (protected-channel mismatch), and $\mathbf q_3=\mathbf e$ (generator inconsistency). Let $\epsilon_i=\|\mathbf P_\perp\mathbf W\mathbf q_i\|_2$ for $i=1,2,3$, $g=\|\mathbf W\mathbf a\|_2^2>0$, and $\kappa=\|\mathbf P_\perp\mathbf W\mathbf a\|_2^2/g$. Then
\begin{equation}
\label{eq:dc_absorb_bound}
\kappa\le
\frac{(\epsilon_1+\epsilon_2+\epsilon_3)^2}{g}.
\end{equation}
Let $n_a=\operatorname{tr}(\boldsymbol\Pi_{\mathcal A})>0$ and $\bar\beta=\operatorname{tr}(\boldsymbol\Pi_{\mathcal A}\mathbf U)/n_a$. An equal-energy random attack, isotropically distributed within the same writable subspace, has mean projected energy $g(1-\bar\beta)$. Therefore, PCGAN has lower projected energy than this random-attack mean exactly when $\kappa<1-\bar\beta$. A sufficient condition is
\begin{equation}
\label{eq:stealth_condition}
\epsilon_1+\epsilon_2+\epsilon_3
<\sqrt{g(1-\bar\beta)}.
\end{equation}
\end{theorem}
\begin{IEEEproof}
Since $\mathbf a=\mathbf H\mathbf T\mathbf u+\sum_{i=1}^{3}\mathbf q_i$ and $\mathbf P_\perp\mathbf W\mathbf H=\mathbf0$, the triangle inequality proves~\eqref{eq:dc_absorb_bound}. For an isotropic random attack $\mathbf a_r$, $\mathbb E_r[(\mathbf W\mathbf a_r)(\mathbf W\mathbf a_r)^\top]=(g/n_a)\boldsymbol\Pi_{\mathcal A}$. Taking the projected-energy trace gives $g(1-\bar\beta)$, establishing~\eqref{eq:stealth_condition}.
\end{IEEEproof}

Theorem~\ref{thm:random} identifies learned-model mismatch, protected-channel responses, and generator inconsistency as factors that limit this advantage. Reducing these errors can decrease residual visibility relative to random attacks, although BDD evasion is not guaranteed.

\begin{corollary}[Ideal stealth and nonlinear AC-WLS bound]
\label{cor:exact}
If $\mathbf E\mathbf u=\mathbf0$, $(\mathbf I_m-\boldsymbol\Pi_{\mathcal A})\mathbf H\mathbf T\mathbf u=\mathbf0$, and $\mathbf e=\mathbf0$, then $\kappa=0$ and $J_\ell(\mathbf a)=J_0$.

More generally, let $\Delta\mathbf x=\mathbf T\mathbf u$, let $\mathcal X$ be the admissible state set containing $\widehat{\mathbf x}+\Delta\mathbf x$, and suppose the whitened AC Jacobian is $L$-Lipschitz along this displacement. The globally minimized nonlinear WLS objective $J_*(\mathbf m+\mathbf a)=\inf_{\mathbf x\in\mathcal X}J_w(\mathbf m+\mathbf a,\mathbf x)$ satisfies
\begin{equation}
\label{eq:nonlinear_bound}
\sqrt{J_*(\mathbf m+\mathbf a)}
\le{}\sqrt{J_0}
+\|\mathbf W(\mathbf D\mathbf u+\mathbf e)\|_2+\frac L2\|\mathbf T\mathbf u\|_2^2,
\end{equation}
where $\mathbf D=\boldsymbol\Pi_{\mathcal A}\mathbf E-(\mathbf I_m-\boldsymbol\Pi_{\mathcal A})\mathbf H\mathbf T$.
\end{corollary}
\begin{IEEEproof}
Under the ideal conditions, $\mathbf a=\mathbf H\Delta\mathbf x$ and its projected residual vanishes. Otherwise, Taylor's theorem bounds the whitened nonlinear remainder by $(L/2)\|\Delta\mathbf x\|_2^2$. Evaluating the attacked residual at $\widehat{\mathbf x}+\Delta\mathbf x$ and applying the triangle inequality proves~\eqref{eq:nonlinear_bound}.
\end{IEEEproof}

The corollary establishes exact linearized stealth under ideal conditions and shows how model mismatch and nonlinear effects constrain it. When the bound in~\eqref{eq:nonlinear_bound} is below $\sqrt{\tau_0}$, the globally minimized residual passes chi-square BDD. This result does not ensure iterative WLS convergence; attack-induced nonconvergence may instead disrupt estimation and trigger operational alarms.

\section{Attack and Detection Benchmarks}
\label{sec:benchmarks}

To assess whether physics guidance improves stealth under restricted access, PCGAN is compared with four conventional and three generative attack baselines. All eight attacks use their permitted sensor channels and are evaluated against the same ten machine-learning (ML) detectors. Conventional attacks alter compromised injection measurements, while generative baselines operate on the writable channels specified by their adaptations.

\textbf{A1}. \emph{Random Attack (RA):} RA applies a bounded multiplicative perturbation to compromised injections \cite{takiddin2024spatiotemporal}, $z_{j,t}^{a}=(1+\alpha_t)z_{j,t}$, where $\mathcal U$ denotes the uniform distribution, $\alpha_r>0$ is the perturbation bound, and $\alpha_t\sim\mathcal U[-\alpha_r,\alpha_r]$ is shared across writable channels at each time step. RA uses neither grid physics nor detector feedback.

\textbf{A2}. \emph{General Attack (GA):} GA scales the injected error to the historical variability of each channel \cite{takiddin2023generalized}, $z_j^{a}=z_j+(-1)^{b_j}\alpha_g\gamma_j\Delta_j$, 
where $b_j\in\{0,1\}$ determines the direction, $\alpha_g>0$ specifies the magnitude, $\gamma_j\sim\mathcal U[0,1]$, and $\Delta_j$ denotes the training range of channel $j$.

\textbf{A3}. \emph{Basic Iterative Method (BIM):} BIM is a projected sign-gradient attack \cite{takiddin2024spatiotemporal} against a surrogate detector the attacker trains on its own readable history and its RA/GA/LRA samples. In standardized coordinates, the compromised injection vector is updated as $\tilde{\mathbf y}_t^{(l+1)}=\mathcal P_{\epsilon}[\tilde{\mathbf y}_t^{(l)}-\alpha_{s}\operatorname{sgn}(\nabla\mathcal L_{c}(\tilde{\mathbf y}_t^{(l)}))]$ from $\tilde{\mathbf y}_t^{(0)}=\tilde{\mathbf y}_t$, where $\mathcal L_{c}$ is the surrogate's loss against the benign label, $\alpha_s>0$ is the step size, and $\mathcal P_{\epsilon}[\cdot]$ projects onto the radius-$\epsilon$ neighborhood of the benign vector under the attack mask. BIM uses the gradients of its own surrogate but not those of the evaluation detectors.

\textbf{A4}. \emph{Load Redistribution Attack (LRA):} LRA redistributes reported active power among the compromised monitored buses while preserving the aggregate \cite{elnour2026joint}, $z_i^{a}=z_i+\Delta z_i$ with $\sum_{i\in\mathcal A}\Delta z_i=0$ and $\max_{i:|z_i|>0}|\Delta z_i/z_i|\le\tau$, where $z_i$ is observed nonzero active power and $\tau$ is the load-shift limit. Zero-injection channels are unchanged. With generation fixed, the perturbation represents a reported-load change $\Delta P_i^{\mathrm d}=-\Delta z_i$.

\textbf{A5}. \emph{LSTMAE-GAN:} LSTMAE-GAN follows the data-driven approach of Narang and Bag~\cite{narang2024physical}. An LSTM autoencoder $\mathcal M(\cdot)$ is trained to reconstruct benign temporal windows and then frozen as a state-estimator mimic. The generator maps an observed measurement, rather than a random prior, to a replacement; the generator and discriminator follow the published convolutional construction with linear adapters for the variable number of channels in the attack region, and benign training samples are split into disjoint pools for generator inputs and discriminator references. Let $\mathbf y_t^{p}$ and $\mathbf y_t^{q}$ be independently drawn normalized source and reference measurements, $\mathbf y_t^{l}=\mathcal G_l(\mathbf y_t^{p})$ the generated replacement clipped to the training range, $\mathbf s_t^{l}$ the temporal window with $\mathbf y_t^{l}$ at its latest position, $\mathcal D_l$ the discriminator, and $n_l$ the number of replaced coordinates. The generator minimizes
\begin{multline}
\label{eq:lstmaegan}
L_l=\ell_l\big(\mathcal D_l(\mathbf y_t^{l})\big)
+\frac{\mu_a}{n_l}\|[\mathcal M(\mathbf s_t^{l})]_t-\mathbf y_t^{l}\|_2^2\\
+\frac{\mu_2}{n_l}\|\mathbf y_t^{l}-\mathbf y_t^{q}\|_2^2
+\frac{\mu_1}{n_l}\|\mathbf y_t^{l}-\mathbf y_t^{p}\|_1,
\end{multline}
where $\ell_l$ is the adversarial binary cross-entropy, $[\mathcal M(\mathbf s_t^{l})]_t$ is the LSTMAE reconstruction at the current time, and $\mu_a,\mu_1,\mu_2>0$ are weights. The quadratic terms enforce reconstruction consistency and reference proximity; the absolute term encourages sparse changes. Boundary line flows readable by the attacker are subtracted from the regional injections before training and added back to the generated output, so that only measurements inside the attack region change while generator-bus injections and boundary flows are left intact.

\textbf{A6}. \emph{Conditional GAN FDI (FDICGAN):} FDICGAN follows Nawaz \emph{et al.} \cite{nawaz2025gan}, who condition both networks on a discrete time-slot label $y_t$, here the clock hour taken from public timestamps, so that offline samples match the demand pattern of that slot. A convolutional generator $\mathcal G_{c}$ maps Gaussian noise $\boldsymbol\xi^{c}$ and the label to a normalized measurement vector, and a convolutional discriminator $\mathcal D_{c}$ is trained with the conditional minimax objective $\min_{\mathcal G_{c}}\max_{\mathcal D_{c}}\mathbb E[\log\mathcal D_{c}(\mathbf z_t\mid y_t)]+\mathbb E[\log(1-\mathcal D_{c}(\mathcal G_{c}(\boldsymbol\xi^{c}\mid y_t)\mid y_t))]$. The attack is the denormalized sample minus the observed measurement on the compromised channels, $\mathbf a_t=\boldsymbol\Pi_{\mathcal A}[\mathcal G_{c}(\boldsymbol\xi^{c}\mid y_t)-\mathbf z_t]$. FDICGAN uses no autoencoder, physics model, or detector feedback, and its generator does not see the current measurement.

\textbf{A7}. \emph{Representation-Learning GAN (RL-iGAN):} RL-iGAN follows Liu \emph{et al.} \cite{liu2026rlgan} in its cross-entropy variant. A decomposable autoencoder with encoder $\mathcal E(\cdot)$ and decoder $\mathcal F(\cdot)$ is trained on benign measurements by minimizing $\mathbb E\|\mathbf z_t-\mathcal F(\mathcal E(\mathbf z_t))\|_2^2$, and the frozen encoder output is rescaled onto $[-1,1]$ by an invertible affine map $\mathcal R_r(\cdot)$, giving latent codes $\boldsymbol\zeta_t=\mathcal R_{r}(\mathcal E(\mathbf z_t))$. A latent generator $\mathcal G_{r}$ and discriminator $\mathcal D_{r}$ are trained with the original minimax objective $\min_{\mathcal G_{r}}\max_{\mathcal D_{r}}\mathbb E[\log\mathcal D_{r}(\boldsymbol\zeta_t)]+\mathbb E[\log(1-\mathcal D_{r}(\mathcal G_{r}(\boldsymbol\xi^{r})))]$ so that a Gaussian prior $\boldsymbol\xi^{r}$ is mapped to codes distributed as $\boldsymbol\zeta_t$. The attack is the decoded sample minus the observed measurement on the compromised channels, $\mathbf a_t=\boldsymbol\Pi_{\mathcal A}[\mathcal F(\mathcal R_{r}^{-1}(\mathcal G_{r}(\boldsymbol\xi^{r})))-\mathbf z_t]$. RL-iGAN is offline and grid-agnostic, requiring no physics model, estimator surrogate, or detector feedback, but, unlike PCGAN and LSTMAE-GAN, its generator ignores the current measurement and context.

Table~\ref{tab:attacker_knowledge} distinguishes the information needed by each attack; the preceding threat model defines which sensor channels each attacker may observe and modify. RA, GA, and BIM alter compromised $P$/$Q$ injections and LRA alters compromised $P$ injections, leaving measured voltages and branch flows unchanged. LSTMAE-GAN modifies the permitted regional measurements while preserving protected boundary channels, whereas FDICGAN, RL-iGAN, and PCGAN are trained only on channels owned by compromised sensors and operate on those writable channels.

\begin{table}[!t]
\centering
\caption{Information required for attack construction.}
\label{tab:attacker_knowledge}
\scriptsize
\setlength{\tabcolsep}{2.1pt}
\renewcommand{\arraystretch}{1.05}
\begin{tabular}{@{}p{0.24\columnwidth} p{0.74\columnwidth}@{}}
\toprule
\textbf{Attack} & \textbf{Required information} \\
\midrule
RA & Current $P$/$Q$ injections and perturbation bound. \\
GA & Historical $P$/$Q$ ranges and perturbation scale. \\
BIM & Readable sensor history, self-trained detector surrogate, and perturbation bound. \\
LRA & Current $P$ injections and zero-sum constraint. \\
LSTMAE-GAN & Historical sensor measurements, local target-bus connectivity, and trained LSTMAE mimic. \\
FDICGAN & Historical sensor measurements and public time labels. \\
RL-iGAN & Historical sensor measurements and learned latent distribution. \\
PCGAN & Historical sensor measurements. \\
\bottomrule
\end{tabular}
\end{table}

\emph{Detection models.} We evaluate ten detectors, comprising a graph convolutional network (GCN), graph diffusion network (GDN), graph transformer network (GTN), their physics-informed (PI) variants (PI-GCN, PI-GDN, PI-GTN), random forest (RF), LSTM, extreme gradient boosting (XGBoost), and a graph attention network (GAT). The graph models retain bus identity and connectivity; physics-informed versions additionally use detector-side AC-consistency residuals~\cite{elnour2026joint}. RF and XGBoost use bus-resolved temporal features, LSTM models measurement sequences, and GAT combines graph attention with temporal aggregation. Detectors are trained on labeled simulated benign and attacked samples; their decision thresholds are calibrated on benign validation data to a 5\% false-positive rate (FPR), with test FPR measured independently. To assess residual-based stealth alongside ML evasion, we also evaluate nominal 95\% AC-WLS chi-square BDD, a 5\%-validation-FPR calibrated chi-square test, and leverage-corrected largest-normalized-residual (LNR) BDD. No defender parameters or test labels are available to PCGAN. The conventional and physics-informed graph detectors are evaluated under the same compromised-channel masks, allowing comparisons under identical sensor budgets. PI detectors use the defender's physics features only; these features are not exposed to the attack generator. The LNR test normalizes individual WLS residuals by their estimated leverage-adjusted variance, complementing the aggregate residual-energy test.

\section{Experimental Methodology}
\label{sec:methodology}
\subsection{Spatiotemporal Load Generation}

We conducted the evaluation on the IEEE 300-bus system, comprising 300 buses, 69 generators, and 411 branches, using observed hourly demand from PJM Interconnection, a regional transmission organization in the United States, between August 2025 and July 2026. Let $D(t)$ denote aggregate demand, with mean $\overline D$, and define $\alpha(t)=D(t)/\overline D$. To reserve 5\% of   generation capacity, a constant scaling factor was applied:
\begin{equation}
\label{eq:load_scale}
\omega=\min\left(1,\frac{0.95P_G^m-P_c}
{P_b\max_t\alpha(t)}\right),
\end{equation}
where $P_G^m$ is the total maximum online generation capacity, $P_b$ is the aggregate positive base-case demand subject to scaling, and $P_c$ accounts for fixed base-case power contributions, including nominal shunts. Since $\omega$ is constant, the relative hourly variations in the PJM profile are preserved. Generation limits, transmission losses, and AC feasibility were subsequently addressed through economic dispatch and AC power flow.

To introduce spatial variability, load buses were grouped into six electrical zones with synthetic residential, commercial, shift-industrial, and continuous-industrial demand profiles. At bus $i$, the profile is $s_i(t)=\sum_c\beta_{i,c}s_{z(i),c}(t)$, where $\beta_{i,c}$ is the proportion of load type $c$, $z(i)$ identifies the electrical zone, and $\sum_c\beta_{i,c}=1$. The profiles incorporated hourly and weekday variations, with residential evening peaks, reduced nighttime commercial and shift-industrial demand, and relatively stable continuous-industrial demand.

Spatially and temporally correlated variations were modeled using a first-order autoregressive process:
\begin{equation}
\label{eq:zone_innov}
\begin{aligned}
\eta_i(t)={}&\rho\eta_i(t-1)
+\sigma_\eta\sqrt{1-\rho^2}\\
&\times\left[\sqrt{0.60}\,\xi_{z(i)}(t)
+\sqrt{0.40}\,\xi_i(t)\right],
\end{aligned}
\end{equation}
where $\xi_{z(i)}(t)$ and $\xi_i(t)$ are independent standard Gaussian innovations shared within each zone and specific to each bus, respectively. We set $\rho=0.85$ and $\sigma_\eta=0.05$. Defining $r_i(t)=s_i(t)e^{\eta_i(t)}$ and $w_i=P_{D,i}^{0}/P_b$, the normalized spatial factor is $f_i(t)=r_i(t)/\sum_jw_jr_j(t)$, where the sum covers positive-load buses. Their active and reactive demands are
\begin{equation}
\label{eq:spatial_demand}
\begin{aligned}
P_{D,i}(t)&=\omega P_{D,i}^{0}\alpha(t)f_i(t),\\
Q_{D,i}(t)&=\omega Q_{D,i}^{0}\alpha(t)f_i(t),
\end{aligned}
\end{equation}
where $P_{D,i}^{0}$ and $Q_{D,i}^{0}$ are the base-case demands. This normalization preserves aggregate active demand and each bus's power factor. Finally, quadratic-cost economic dispatch (ED) and AC power flow generated the benign operating states.

To represent sparse instrumentation, mixed-integer optimization selected a minimum-cardinality one-hop-cover deployment with load-bus and criticality tie-breaks~\cite{HosseiniDolatabadi2026,Mabaning2026}, placing PMU/RTU pairs at 87 buses. Their measurements include voltage magnitudes, angles, injections, and incident branch-end flows. The nested compromise sets contain 9, 26, 52, and 87 sensors: S1 (10\%), S2 (30\%), S3 (60\%), and S4 (100\%) of the deployed sensors. Attacks modify only owned measurement channels; at S4 the attacker can replace the entire collected vector, but not unmeasured states.  The optimized deployment was fixed across all scenarios, and compromised sensors were selected in nested order. Holding the network and meter placement fixed isolates the effect of increasing attacker access. For S1--S3, protected neighboring measurements can expose inconsistencies in altered injections or flows during AC-WLS estimation.

\begin{table*}[!b]
\centering
\caption{Attack detector performance under S1--S4 (10\%, 30\%, 60\%, 100\% compromised sensors). Mean $\pm$ sample SD (\%) over seeds 41--43; ML metrics use 5\% validation FPR. Bold: highest mean evasion for each detector within each scenario.}
\label{tab:detector_all_attacks_pjm_5pct}
\setlength{\tabcolsep}{1.10pt}
\renewcommand{\arraystretch}{0.95}
\scriptsize
\resizebox{\textwidth}{!}{%
\begin{tabular}{@{}ll*{32}{c}@{}}
\toprule
\multirow{2}{*}{\textbf{Detector}} & \multirow{2}{*}{\textbf{Metric}} & \multicolumn{4}{c}{\textbf{RA}} & \multicolumn{4}{c}{\textbf{GA}} & \multicolumn{4}{c}{\textbf{BIM}} & \multicolumn{4}{c}{\textbf{LRA}} & \multicolumn{4}{c}{\textbf{LSTMAE-GAN}} & \multicolumn{4}{c}{\textbf{FDICGAN}} & \multicolumn{4}{c}{\textbf{RL-iGAN}} & \multicolumn{4}{c}{\textbf{PCGAN}} \\
\cmidrule(lr){3-6} \cmidrule(lr){7-10} \cmidrule(lr){11-14} \cmidrule(lr){15-18} \cmidrule(lr){19-22} \cmidrule(lr){23-26} \cmidrule(lr){27-30} \cmidrule(lr){31-34}
 & & \textbf{10\%} & \textbf{30\%} & \textbf{60\%} & \textbf{100\%} & \textbf{10\%} & \textbf{30\%} & \textbf{60\%} & \textbf{100\%} & \textbf{10\%} & \textbf{30\%} & \textbf{60\%} & \textbf{100\%} & \textbf{10\%} & \textbf{30\%} & \textbf{60\%} & \textbf{100\%} & \textbf{10\%} & \textbf{30\%} & \textbf{60\%} & \textbf{100\%} & \textbf{10\%} & \textbf{30\%} & \textbf{60\%} & \textbf{100\%} & \textbf{10\%} & \textbf{30\%} & \textbf{60\%} & \textbf{100\%} & \textbf{10\%} & \textbf{30\%} & \textbf{60\%} & \textbf{100\%} \\
\midrule
\multirow{5}{*}{\textbf{RF}} & Accuracy & 54.1$\pm$0.6 & 56.4$\pm$1.1 & 58.5$\pm$1.7 & 58.1$\pm$1.7 & 59.0$\pm$2.6 & 79.0$\pm$5.0 & 97.2$\pm$0.4 & 97.6$\pm$0.7 & 50.1$\pm$0.3 & 51.3$\pm$0.6 & 65.2$\pm$6.6 & 57.6$\pm$3.0 & 97.3$\pm$0.6 & 98.0$\pm$0.6 & 98.2$\pm$0.6 & 98.2$\pm$0.6 & 96.1$\pm$2.8 & 94.1$\pm$2.1 & 94.2$\pm$2.0 & 95.3$\pm$2.5 & 98.3$\pm$0.7 & 98.3$\pm$0.7 & 98.3$\pm$0.7 & 98.3$\pm$0.7 & 98.3$\pm$0.7 & 98.3$\pm$0.7 & 98.3$\pm$0.7 & 91.4$\pm$2.7 & 52.1$\pm$0.6 & 50.6$\pm$0.1 & 50.8$\pm$0.2 & 51.2$\pm$0.5 \\
  & F1-score & 20.2$\pm$3.3 & 27.0$\pm$4.7 & 32.7$\pm$6.0 & 31.7$\pm$5.8 & 34.0$\pm$8.0 & 74.2$\pm$7.9 & 97.2$\pm$0.4 & 97.7$\pm$0.7 & 6.6$\pm$3.5 & 10.9$\pm$4.3 & 48.1$\pm$14.6 & 30.2$\pm$10.1 & 97.3$\pm$0.5 & 98.0$\pm$0.6 & 98.2$\pm$0.6 & 98.2$\pm$0.6 & 96.0$\pm$3.1 & 94.0$\pm$2.3 & 94.1$\pm$2.1 & 95.2$\pm$2.7 & 98.3$\pm$0.6 & 98.3$\pm$0.7 & 98.3$\pm$0.7 & 98.3$\pm$0.7 & 98.3$\pm$0.7 & 98.3$\pm$0.7 & 98.3$\pm$0.6 & 90.8$\pm$3.2 & 13.6$\pm$1.9 & 8.4$\pm$2.1 & 9.2$\pm$2.7 & 10.5$\pm$1.9 \\
  & Precision & 77.9$\pm$4.5 & 83.1$\pm$3.2 & 86.0$\pm$2.6 & 85.5$\pm$3.3 & 86.5$\pm$2.6 & 94.7$\pm$1.9 & 96.6$\pm$1.3 & 96.7$\pm$1.3 & 49.2$\pm$7.2 & 63.5$\pm$1.0 & 90.7$\pm$1.8 & 84.5$\pm$1.7 & 96.6$\pm$1.3 & 96.7$\pm$1.3 & 96.7$\pm$1.3 & 96.7$\pm$1.3 & 96.6$\pm$1.2 & 96.4$\pm$1.4 & 96.4$\pm$1.4 & 96.5$\pm$1.3 & 96.7$\pm$1.3 & 96.7$\pm$1.3 & 96.7$\pm$1.3 & 96.7$\pm$1.3 & 96.7$\pm$1.3 & 96.7$\pm$1.3 & 96.7$\pm$1.3 & 96.2$\pm$1.4 & 69.6$\pm$7.7 & 58.1$\pm$3.8 & 60.2$\pm$3.4 & 63.6$\pm$8.4 \\
  & Evasion & 88.4$\pm$2.2 & 83.8$\pm$3.4 & 79.6$\pm$4.6 & 80.4$\pm$4.3 & 78.6$\pm$6.5 & 38.6$\pm$10.4 & 2.2$\pm$0.6 & 1.3$\pm$0.5 & \textbf{96.4$\pm$2.0} & 94.0$\pm$2.6 & 66.3$\pm$14.4 & 81.3$\pm$7.2 & 2.0$\pm$0.7 & 0.6$\pm$0.1 & 0.2$\pm$0.2 & 0.2$\pm$0.2 & 4.5$\pm$6.9 & 8.3$\pm$4.4 & 8.1$\pm$4.1 & 5.9$\pm$5.8 & 0.1$\pm$0.1 & 0.0$\pm$0.0 & 0.0$\pm$0.0 & 0.0$\pm$0.0 & 0.0$\pm$0.0 & 0.0$\pm$0.0 & 0.1$\pm$0.1 & 13.8$\pm$6.1 & 92.4$\pm$1.2 & \textbf{95.4$\pm$1.2} & \textbf{95.0$\pm$1.6} & \textbf{94.2$\pm$1.2} \\
  & AUROC & 64.3$\pm$0.4 & 68.1$\pm$0.6 & 71.7$\pm$0.8 & 71.2$\pm$0.8 & 78.3$\pm$2.0 & 93.8$\pm$2.3 & 99.6$\pm$0.1 & 99.8$\pm$0.1 & 52.2$\pm$0.4 & 57.1$\pm$2.5 & 82.0$\pm$5.7 & 74.2$\pm$1.8 & 99.7$\pm$0.1 & 99.9$\pm$0.1 & 100.0$\pm$0.0 & 100.0$\pm$0.0 & 99.3$\pm$1.1 & 98.2$\pm$0.9 & 98.3$\pm$0.8 & 98.8$\pm$0.7 & 100.0$\pm$0.0 & 100.0$\pm$0.0 & 100.0$\pm$0.0 & 100.0$\pm$0.0 & 100.0$\pm$0.0 & 100.0$\pm$0.0 & 100.0$\pm$0.0 & 94.3$\pm$3.3 & 58.3$\pm$2.1 & 53.3$\pm$0.4 & 53.7$\pm$0.2 & 56.3$\pm$1.5 \\
\midrule
\multirow{5}{*}{\textbf{LSTM}} & Accuracy & 65.0$\pm$2.2 & 72.7$\pm$1.7 & 84.3$\pm$0.6 & 83.7$\pm$0.8 & 62.6$\pm$3.3 & 85.7$\pm$3.7 & 93.1$\pm$1.3 & 93.1$\pm$1.2 & 51.8$\pm$1.0 & 59.1$\pm$3.1 & 78.2$\pm$6.3 & 68.3$\pm$5.6 & 91.0$\pm$1.5 & 90.4$\pm$1.3 & 91.6$\pm$1.9 & 92.0$\pm$1.3 & 94.4$\pm$2.6 & 91.4$\pm$2.4 & 91.4$\pm$2.5 & 91.0$\pm$1.8 & 97.1$\pm$1.7 & 97.1$\pm$1.7 & 97.1$\pm$1.7 & 97.1$\pm$1.8 & 97.1$\pm$1.7 & 97.1$\pm$1.7 & 97.1$\pm$1.7 & 86.4$\pm$3.9 & 54.5$\pm$0.2 & 51.8$\pm$0.2 & 52.3$\pm$0.5 & 54.1$\pm$0.8 \\
  & F1-score & 50.3$\pm$6.8 & 65.0$\pm$3.9 & 82.5$\pm$1.3 & 81.7$\pm$1.6 & 45.0$\pm$7.4 & 84.3$\pm$4.1 & 93.0$\pm$1.2 & 93.0$\pm$1.2 & 16.0$\pm$5.3 & 36.3$\pm$10.7 & 73.8$\pm$8.5 & 56.4$\pm$12.1 & 90.7$\pm$1.4 & 90.0$\pm$1.2 & 91.4$\pm$1.8 & 91.9$\pm$1.1 & 94.4$\pm$2.7 & 91.1$\pm$2.9 & 91.1$\pm$2.9 & 90.7$\pm$2.4 & 97.2$\pm$1.6 & 97.2$\pm$1.6 & 97.2$\pm$1.6 & 97.2$\pm$1.7 & 97.2$\pm$1.6 & 97.2$\pm$1.6 & 97.2$\pm$1.6 & 85.1$\pm$4.5 & 24.3$\pm$4.2 & 15.9$\pm$5.4 & 17.6$\pm$5.5 & 23.1$\pm$5.4 \\
  & Precision & 86.9$\pm$4.9 & 90.3$\pm$4.5 & 93.1$\pm$3.5 & 93.0$\pm$3.5 & 84.9$\pm$7.2 & 93.1$\pm$4.2 & 94.2$\pm$3.2 & 94.2$\pm$3.2 & 63.2$\pm$6.9 & 81.4$\pm$4.5 & 91.4$\pm$5.8 & 88.4$\pm$5.0 & 94.0$\pm$3.4 & 93.9$\pm$3.4 & 94.0$\pm$3.4 & 94.1$\pm$3.3 & 94.4$\pm$3.2 & 94.1$\pm$2.9 & 94.1$\pm$2.9 & 94.1$\pm$3.0 & 94.6$\pm$3.1 & 94.6$\pm$3.1 & 94.6$\pm$3.1 & 94.6$\pm$3.1 & 94.6$\pm$3.1 & 94.6$\pm$3.1 & 94.6$\pm$3.1 & 93.3$\pm$4.0 & 73.6$\pm$7.3 & 63.3$\pm$4.6 & 65.9$\pm$5.3 & 72.3$\pm$7.3 \\
  & Evasion & 64.2$\pm$7.2 & 48.9$\pm$5.7 & 25.7$\pm$4.4 & 26.8$\pm$4.7 & 69.1$\pm$6.7 & 22.9$\pm$4.2 & 8.1$\pm$2.0 & 8.1$\pm$2.6 & \textbf{90.7$\pm$3.5} & 76.1$\pm$8.8 & 37.9$\pm$10.3 & 57.6$\pm$12.5 & 12.2$\pm$1.0 & 13.4$\pm$2.0 & 11.0$\pm$0.8 & 10.2$\pm$1.0 & 5.4$\pm$5.4 & 11.4$\pm$7.5 & 11.4$\pm$7.3 & 12.2$\pm$6.7 & 0.1$\pm$0.1 & 0.0$\pm$0.0 & 0.0$\pm$0.0 & 0.0$\pm$0.1 & 0.0$\pm$0.0 & 0.0$\pm$0.1 & 0.1$\pm$0.1 & 21.5$\pm$6.6 & 85.2$\pm$3.5 & \textbf{90.7$\pm$3.8} & \textbf{89.6$\pm$3.9} & \textbf{86.0$\pm$4.2} \\
  & AUROC & 76.5$\pm$2.6 & 84.4$\pm$2.4 & 93.1$\pm$1.2 & 92.7$\pm$1.1 & 73.7$\pm$4.0 & 94.5$\pm$2.6 & 98.1$\pm$0.6 & 98.3$\pm$0.6 & 54.6$\pm$2.9 & 70.8$\pm$3.5 & 89.1$\pm$4.6 & 80.6$\pm$2.9 & 96.6$\pm$0.9 & 96.5$\pm$0.8 & 97.1$\pm$1.3 & 97.4$\pm$0.8 & 98.6$\pm$1.4 & 97.2$\pm$1.6 & 97.3$\pm$1.5 & 96.9$\pm$0.9 & 100.0$\pm$0.0 & 100.0$\pm$0.0 & 100.0$\pm$0.0 & 100.0$\pm$0.0 & 100.0$\pm$0.0 & 100.0$\pm$0.0 & 100.0$\pm$0.0 & 93.7$\pm$3.7 & 59.5$\pm$1.5 & 54.6$\pm$0.6 & 54.9$\pm$1.6 & 57.9$\pm$2.8 \\
\midrule
\multirow{5}{*}{\textbf{XGBoost}} & Accuracy & 62.4$\pm$1.0 & 64.1$\pm$1.5 & 66.9$\pm$1.9 & 66.1$\pm$2.1 & 73.3$\pm$4.5 & 87.3$\pm$4.6 & 96.3$\pm$0.4 & 96.5$\pm$0.5 & 51.1$\pm$1.4 & 54.9$\pm$4.9 & 80.4$\pm$5.0 & 71.0$\pm$3.0 & 96.4$\pm$0.6 & 96.6$\pm$0.5 & 96.7$\pm$0.6 & 96.6$\pm$0.5 & 95.7$\pm$2.0 & 94.0$\pm$1.3 & 94.1$\pm$1.0 & 94.5$\pm$1.7 & 96.8$\pm$0.6 & 96.8$\pm$0.6 & 96.8$\pm$0.6 & 96.8$\pm$0.6 & 96.8$\pm$0.6 & 96.8$\pm$0.5 & 96.8$\pm$0.6 & 91.6$\pm$2.1 & 53.5$\pm$0.8 & 50.8$\pm$0.8 & 51.2$\pm$0.6 & 52.2$\pm$0.4 \\
  & F1-score & 45.3$\pm$2.6 & 49.1$\pm$3.2 & 54.7$\pm$4.0 & 53.0$\pm$4.3 & 66.2$\pm$7.1 & 86.4$\pm$5.6 & 96.4$\pm$0.4 & 96.6$\pm$0.4 & 14.8$\pm$3.2 & 25.9$\pm$12.4 & 77.2$\pm$6.6 & 62.4$\pm$5.5 & 96.5$\pm$0.5 & 96.7$\pm$0.5 & 96.8$\pm$0.6 & 96.7$\pm$0.5 & 95.7$\pm$2.1 & 94.0$\pm$1.4 & 94.1$\pm$1.1 & 94.6$\pm$1.8 & 96.9$\pm$0.6 & 96.9$\pm$0.5 & 96.9$\pm$0.5 & 96.9$\pm$0.5 & 96.9$\pm$0.5 & 96.9$\pm$0.5 & 96.9$\pm$0.6 & 91.3$\pm$2.4 & 22.3$\pm$1.5 & 14.0$\pm$3.0 & 15.1$\pm$1.3 & 18.4$\pm$2.7 \\
  & Precision & 83.1$\pm$2.1 & 84.5$\pm$2.5 & 86.3$\pm$2.1 & 85.8$\pm$2.2 & 89.0$\pm$2.9 & 92.6$\pm$1.9 & 94.0$\pm$1.0 & 94.0$\pm$1.0 & 56.9$\pm$8.8 & 68.8$\pm$12.8 & 91.2$\pm$2.4 & 88.3$\pm$2.0 & 94.0$\pm$1.0 & 94.0$\pm$1.0 & 94.0$\pm$1.0 & 94.0$\pm$1.0 & 93.9$\pm$1.0 & 93.7$\pm$0.9 & 93.7$\pm$0.9 & 93.8$\pm$1.0 & 94.0$\pm$1.0 & 94.0$\pm$1.0 & 94.0$\pm$1.0 & 94.0$\pm$1.0 & 94.0$\pm$1.0 & 94.0$\pm$1.0 & 94.0$\pm$1.0 & 93.4$\pm$1.1 & 67.8$\pm$4.2 & 55.6$\pm$5.6 & 57.9$\pm$4.4 & 63.0$\pm$1.3 \\
  & Evasion & 68.8$\pm$2.5 & 65.3$\pm$3.0 & 59.9$\pm$4.1 & 61.5$\pm$4.4 & 47.1$\pm$8.1 & 18.9$\pm$8.3 & 1.0$\pm$0.3 & 0.7$\pm$0.6 & \textbf{91.5$\pm$1.9} & 83.8$\pm$8.9 & 32.8$\pm$8.8 & 51.6$\pm$6.2 & 0.9$\pm$0.1 & 0.5$\pm$0.1 & 0.2$\pm$0.0 & 0.4$\pm$0.1 & 2.3$\pm$3.8 & 5.6$\pm$3.4 & 5.5$\pm$2.9 & 4.6$\pm$3.7 & 0.1$\pm$0.1 & 0.0$\pm$0.0 & 0.0$\pm$0.0 & 0.0$\pm$0.0 & 0.0$\pm$0.1 & 0.0$\pm$0.1 & 0.1$\pm$0.1 & 10.5$\pm$4.1 & 86.6$\pm$1.0 & \textbf{92.0$\pm$1.9} & \textbf{91.3$\pm$0.8} & \textbf{89.2$\pm$1.8} \\
  & AUROC & 72.1$\pm$0.4 & 74.5$\pm$1.2 & 77.7$\pm$1.7 & 76.9$\pm$1.7 & 86.1$\pm$3.5 & 95.2$\pm$2.7 & 99.7$\pm$0.1 & 99.8$\pm$0.1 & 55.8$\pm$4.7 & 64.9$\pm$8.6 & 90.1$\pm$4.4 & 83.9$\pm$1.4 & 99.8$\pm$0.1 & 99.8$\pm$0.0 & 99.9$\pm$0.0 & 99.9$\pm$0.0 & 99.4$\pm$0.9 & 98.4$\pm$0.9 & 98.5$\pm$0.8 & 98.7$\pm$0.9 & 100.0$\pm$0.0 & 100.0$\pm$0.0 & 100.0$\pm$0.0 & 100.0$\pm$0.0 & 100.0$\pm$0.0 & 100.0$\pm$0.0 & 100.0$\pm$0.0 & 97.1$\pm$0.9 & 60.0$\pm$1.7 & 53.9$\pm$1.1 & 55.0$\pm$0.7 & 56.8$\pm$1.4 \\
\midrule
\multirow{5}{*}{\textbf{GAT}} & Accuracy & 54.0$\pm$3.1 & 55.4$\pm$3.3 & 57.2$\pm$3.6 & 55.7$\pm$2.4 & 61.7$\pm$7.0 & 75.6$\pm$7.1 & 85.0$\pm$4.2 & 89.4$\pm$4.5 & 51.4$\pm$0.8 & 53.3$\pm$3.6 & 58.0$\pm$2.6 & 54.9$\pm$0.7 & 91.9$\pm$1.6 & 92.5$\pm$2.3 & 95.7$\pm$0.6 & 96.2$\pm$0.2 & 95.4$\pm$1.2 & 93.4$\pm$1.4 & 93.6$\pm$1.5 & 93.9$\pm$2.2 & 97.3$\pm$1.1 & 97.3$\pm$1.1 & 97.3$\pm$1.1 & 97.3$\pm$1.1 & 97.3$\pm$1.1 & 97.3$\pm$1.1 & 97.3$\pm$1.1 & 90.9$\pm$4.2 & 55.0$\pm$1.3 & 52.6$\pm$0.7 & 52.0$\pm$0.2 & 56.7$\pm$2.1 \\
  & F1-score & 21.9$\pm$11.7 & 25.9$\pm$11.8 & 30.8$\pm$12.3 & 27.0$\pm$9.5 & 41.4$\pm$17.0 & 69.1$\pm$11.5 & 83.1$\pm$6.1 & 88.5$\pm$5.7 & 14.2$\pm$6.0 & 19.6$\pm$13.3 & 33.4$\pm$8.0 & 25.0$\pm$4.7 & 91.6$\pm$1.9 & 92.3$\pm$2.7 & 95.7$\pm$0.6 & 96.3$\pm$0.2 & 95.4$\pm$1.4 & 93.3$\pm$1.6 & 93.5$\pm$1.7 & 93.8$\pm$2.5 & 97.4$\pm$1.0 & 97.3$\pm$1.1 & 97.4$\pm$1.0 & 97.4$\pm$1.0 & 97.4$\pm$1.0 & 97.3$\pm$1.1 & 97.4$\pm$1.0 & 90.4$\pm$4.9 & 25.4$\pm$4.2 & 18.4$\pm$1.4 & 16.4$\pm$4.0 & 30.2$\pm$4.1 \\
  & Precision & 69.4$\pm$4.5 & 74.0$\pm$2.6 & 77.9$\pm$1.6 & 75.3$\pm$1.5 & 83.4$\pm$3.9 & 91.3$\pm$2.2 & 93.4$\pm$1.8 & 94.1$\pm$1.7 & 59.4$\pm$2.0 & 65.4$\pm$7.8 & 79.8$\pm$5.0 & 74.3$\pm$3.5 & 94.3$\pm$1.9 & 94.4$\pm$1.8 & 94.7$\pm$1.9 & 94.8$\pm$1.9 & 94.7$\pm$1.8 & 94.5$\pm$2.0 & 94.5$\pm$1.9 & 94.6$\pm$1.8 & 94.9$\pm$2.0 & 94.8$\pm$2.0 & 94.9$\pm$2.0 & 94.9$\pm$2.0 & 94.9$\pm$2.0 & 94.8$\pm$2.0 & 94.9$\pm$2.0 & 94.2$\pm$1.9 & 74.2$\pm$6.9 & 67.1$\pm$7.5 & 64.2$\pm$3.1 & 77.6$\pm$8.1 \\
  & Evasion & 86.6$\pm$8.3 & 83.8$\pm$8.9 & 80.2$\pm$9.5 & 83.2$\pm$7.0 & 71.2$\pm$16.1 & 43.3$\pm$16.0 & 24.5$\pm$10.5 & 15.8$\pm$10.9 & \textbf{91.8$\pm$3.9} & 88.0$\pm$9.3 & 78.6$\pm$6.7 & \textbf{84.8$\pm$3.6} & 10.8$\pm$5.3 & 9.5$\pm$6.8 & 3.2$\pm$2.3 & 2.1$\pm$1.7 & 3.7$\pm$4.4 & 7.7$\pm$4.2 & 7.4$\pm$4.6 & 6.8$\pm$6.2 & 0.0$\pm$0.0 & 0.0$\pm$0.0 & 0.0$\pm$0.0 & 0.0$\pm$0.0 & 0.0$\pm$0.0 & 0.0$\pm$0.0 & 0.0$\pm$0.0 & 12.7$\pm$9.6 & 84.6$\pm$3.1 & \textbf{89.3$\pm$1.1} & \textbf{90.5$\pm$2.7} & 81.2$\pm$2.8 \\
  & AUROC & 62.3$\pm$3.3 & 64.4$\pm$3.7 & 68.9$\pm$3.7 & 66.6$\pm$3.0 & 73.2$\pm$6.3 & 87.3$\pm$4.9 & 94.3$\pm$1.8 & 96.4$\pm$1.9 & 54.9$\pm$3.3 & 58.6$\pm$5.7 & 69.0$\pm$5.5 & 63.4$\pm$2.5 & 97.4$\pm$0.8 & 98.0$\pm$1.0 & 99.4$\pm$0.3 & 99.6$\pm$0.2 & 99.1$\pm$1.1 & 98.1$\pm$0.6 & 98.2$\pm$0.6 & 98.4$\pm$1.2 & 100.0$\pm$0.0 & 100.0$\pm$0.0 & 100.0$\pm$0.0 & 100.0$\pm$0.0 & 100.0$\pm$0.0 & 100.0$\pm$0.0 & 100.0$\pm$0.0 & 96.6$\pm$2.8 & 59.7$\pm$2.7 & 55.4$\pm$1.6 & 55.7$\pm$1.1 & 62.8$\pm$4.2 \\
\midrule
\multirow{5}{*}{\textbf{GCN}} & Accuracy & 51.9$\pm$0.8 & 52.0$\pm$0.8 & 53.9$\pm$0.9 & 53.0$\pm$1.0 & 55.3$\pm$2.8 & 65.7$\pm$5.8 & 77.9$\pm$3.7 & 86.2$\pm$4.3 & 50.4$\pm$0.3 & 51.7$\pm$0.9 & 55.2$\pm$2.5 & 52.9$\pm$1.2 & 86.9$\pm$2.0 & 87.5$\pm$2.6 & 95.4$\pm$1.2 & 95.7$\pm$1.5 & 94.8$\pm$1.7 & 91.2$\pm$3.1 & 91.6$\pm$2.7 & 91.8$\pm$3.4 & 97.6$\pm$1.4 & 97.6$\pm$1.4 & 97.6$\pm$1.4 & 97.6$\pm$1.4 & 97.6$\pm$1.4 & 97.6$\pm$1.4 & 97.6$\pm$1.4 & 88.0$\pm$2.7 & 53.9$\pm$0.4 & 52.5$\pm$1.0 & 52.4$\pm$0.8 & 58.1$\pm$2.3 \\
  & F1-score & 15.1$\pm$6.4 & 15.2$\pm$6.6 & 21.4$\pm$6.5 & 18.4$\pm$6.7 & 25.1$\pm$10.7 & 50.6$\pm$12.4 & 73.0$\pm$6.2 & 84.7$\pm$5.5 & 10.2$\pm$5.1 & 14.5$\pm$5.8 & 24.8$\pm$9.0 & 18.5$\pm$0.7 & 85.7$\pm$2.7 & 86.4$\pm$3.4 & 95.5$\pm$1.2 & 95.8$\pm$1.5 & 94.7$\pm$1.9 & 90.7$\pm$3.5 & 91.3$\pm$3.1 & 91.4$\pm$4.2 & 97.6$\pm$1.3 & 97.6$\pm$1.3 & 97.6$\pm$1.3 & 97.6$\pm$1.3 & 97.6$\pm$1.3 & 97.6$\pm$1.3 & 97.6$\pm$1.3 & 87.0$\pm$3.3 & 21.4$\pm$2.6 & 17.1$\pm$2.9 & 16.6$\pm$4.0 & 33.4$\pm$3.7 \\
  & Precision & 65.5$\pm$4.6 & 65.7$\pm$4.4 & 74.1$\pm$6.4 & 70.5$\pm$6.0 & 76.8$\pm$6.8 & 88.1$\pm$6.7 & 92.9$\pm$3.4 & 94.1$\pm$3.2 & 54.7$\pm$2.5 & 64.9$\pm$8.1 & 76.6$\pm$9.4 & 70.5$\pm$13.0 & 94.4$\pm$2.9 & 94.5$\pm$2.8 & 95.2$\pm$2.6 & 95.2$\pm$2.6 & 95.2$\pm$2.5 & 94.8$\pm$2.8 & 94.9$\pm$2.7 & 95.0$\pm$2.5 & 95.4$\pm$2.5 & 95.4$\pm$2.5 & 95.4$\pm$2.5 & 95.4$\pm$2.5 & 95.4$\pm$2.5 & 95.4$\pm$2.5 & 95.4$\pm$2.5 & 94.4$\pm$2.9 & 74.0$\pm$10.1 & 68.9$\pm$12.2 & 68.5$\pm$10.7 & 81.5$\pm$10.3 \\
  & Evasion & 91.3$\pm$4.0 & 91.2$\pm$4.2 & 87.3$\pm$4.4 & 89.2$\pm$4.4 & 84.6$\pm$7.8 & 63.7$\pm$12.6 & 39.4$\pm$9.6 & 22.8$\pm$8.3 & \textbf{94.3$\pm$3.0} & \textbf{91.7$\pm$3.7} & 84.8$\pm$6.7 & \textbf{89.3$\pm$0.6} & 21.3$\pm$6.0 & 20.1$\pm$7.4 & 4.3$\pm$1.5 & 3.6$\pm$1.6 & 5.5$\pm$5.9 & 12.8$\pm$7.2 & 11.9$\pm$6.6 & 11.5$\pm$9.5 & 0.0$\pm$0.0 & 0.0$\pm$0.0 & 0.0$\pm$0.0 & 0.0$\pm$0.0 & 0.0$\pm$0.0 & 0.0$\pm$0.0 & 0.0$\pm$0.0 & 19.2$\pm$6.4 & 87.4$\pm$2.0 & 90.2$\pm$2.0 & \textbf{90.4$\pm$2.6} & 79.0$\pm$2.4 \\
  & AUROC & 57.3$\pm$1.4 & 57.6$\pm$2.7 & 62.2$\pm$1.4 & 60.4$\pm$0.7 & 64.4$\pm$5.5 & 78.4$\pm$6.9 & 90.2$\pm$2.5 & 94.4$\pm$3.3 & 52.5$\pm$1.7 & 55.7$\pm$3.5 & 65.2$\pm$8.0 & 61.1$\pm$4.8 & 94.7$\pm$0.9 & 95.0$\pm$1.6 & 99.0$\pm$0.5 & 99.2$\pm$0.5 & 98.7$\pm$1.3 & 96.3$\pm$2.0 & 96.5$\pm$1.7 & 96.9$\pm$1.9 & 100.0$\pm$0.0 & 100.0$\pm$0.0 & 100.0$\pm$0.0 & 100.0$\pm$0.0 & 100.0$\pm$0.0 & 100.0$\pm$0.0 & 100.0$\pm$0.0 & 94.0$\pm$3.6 & 57.8$\pm$1.9 & 55.6$\pm$0.5 & 55.9$\pm$0.8 & 64.8$\pm$4.1 \\
\midrule
\multirow{5}{*}{\textbf{GDN}} & Accuracy & 51.8$\pm$0.6 & 51.9$\pm$0.5 & 54.7$\pm$1.0 & 53.5$\pm$0.6 & 55.3$\pm$2.0 & 64.5$\pm$5.9 & 77.5$\pm$7.6 & 83.4$\pm$7.3 & 50.3$\pm$0.2 & 51.9$\pm$0.8 & 55.9$\pm$4.3 & 54.0$\pm$2.7 & 87.5$\pm$0.5 & 89.0$\pm$1.7 & 95.6$\pm$1.0 & 95.6$\pm$0.8 & 94.4$\pm$3.1 & 90.0$\pm$3.8 & 90.3$\pm$3.5 & 91.7$\pm$3.5 & 97.0$\pm$0.5 & 97.0$\pm$0.5 & 97.0$\pm$0.5 & 97.0$\pm$0.5 & 97.0$\pm$0.5 & 97.0$\pm$0.5 & 97.0$\pm$0.5 & 87.8$\pm$1.8 & 54.2$\pm$1.0 & 52.6$\pm$0.8 & 52.5$\pm$0.8 & 59.1$\pm$3.0 \\
  & F1-score & 16.7$\pm$2.5 & 17.0$\pm$1.4 & 25.4$\pm$1.8 & 22.0$\pm$1.3 & 27.2$\pm$5.1 & 48.9$\pm$13.1 & 72.4$\pm$11.4 & 80.9$\pm$10.0 & 11.9$\pm$1.8 & 16.9$\pm$2.7 & 28.3$\pm$11.8 & 23.1$\pm$8.2 & 86.6$\pm$0.5 & 88.4$\pm$1.9 & 95.7$\pm$1.0 & 95.7$\pm$0.8 & 94.4$\pm$3.2 & 89.4$\pm$4.4 & 89.8$\pm$4.0 & 91.4$\pm$3.9 & 97.0$\pm$0.4 & 97.0$\pm$0.4 & 97.0$\pm$0.4 & 97.0$\pm$0.4 & 97.0$\pm$0.4 & 97.0$\pm$0.4 & 97.0$\pm$0.4 & 87.0$\pm$2.1 & 24.0$\pm$3.5 & 19.2$\pm$3.7 & 18.9$\pm$3.7 & 36.9$\pm$8.1 \\
  & Precision & 61.3$\pm$2.5 & 62.0$\pm$3.3 & 71.7$\pm$4.6 & 68.4$\pm$3.6 & 72.9$\pm$6.3 & 84.2$\pm$6.0 & 90.6$\pm$2.9 & 92.1$\pm$2.2 & 52.6$\pm$1.1 & 61.6$\pm$4.2 & 71.8$\pm$12.8 & 68.3$\pm$9.0 & 93.0$\pm$1.0 & 93.3$\pm$1.0 & 94.1$\pm$0.9 & 94.1$\pm$0.9 & 93.9$\pm$1.2 & 93.4$\pm$0.8 & 93.5$\pm$0.8 & 93.6$\pm$1.3 & 94.3$\pm$0.8 & 94.3$\pm$0.8 & 94.3$\pm$0.8 & 94.3$\pm$0.8 & 94.3$\pm$0.8 & 94.3$\pm$0.8 & 94.3$\pm$0.8 & 93.1$\pm$1.1 & 70.4$\pm$2.7 & 64.8$\pm$1.7 & 64.4$\pm$1.7 & 79.6$\pm$2.2 \\
  & Evasion & 90.3$\pm$1.6 & 90.1$\pm$1.0 & 84.6$\pm$1.1 & \textbf{86.9$\pm$0.9} & 83.3$\pm$3.5 & 65.0$\pm$11.7 & 38.9$\pm$14.6 & 27.2$\pm$14.2 & \textbf{93.3$\pm$1.1} & \textbf{90.2$\pm$1.8} & 82.2$\pm$8.2 & 86.0$\pm$5.7 & 18.9$\pm$1.2 & 15.9$\pm$3.0 & 2.7$\pm$1.6 & 2.8$\pm$1.2 & 5.1$\pm$5.3 & 14.0$\pm$8.1 & 13.3$\pm$7.5 & 10.5$\pm$6.4 & 0.0$\pm$0.0 & 0.0$\pm$0.0 & 0.0$\pm$0.0 & 0.0$\pm$0.0 & 0.0$\pm$0.0 & 0.0$\pm$0.0 & 0.0$\pm$0.0 & 18.3$\pm$3.1 & 85.5$\pm$2.5 & 88.7$\pm$2.5 & \textbf{88.9$\pm$2.5} & 75.8$\pm$6.8 \\
  & AUROC & 57.1$\pm$1.2 & 58.1$\pm$2.3 & 63.6$\pm$2.3 & 61.7$\pm$1.6 & 63.3$\pm$4.2 & 76.3$\pm$6.5 & 88.1$\pm$5.1 & 92.4$\pm$4.9 & 52.0$\pm$1.4 & 54.7$\pm$2.9 & 65.3$\pm$10.5 & 61.5$\pm$6.0 & 94.5$\pm$0.2 & 95.7$\pm$1.0 & 99.2$\pm$0.5 & 99.3$\pm$0.3 & 98.3$\pm$2.0 & 95.4$\pm$2.5 & 95.6$\pm$2.2 & 96.8$\pm$2.4 & 100.0$\pm$0.0 & 100.0$\pm$0.0 & 100.0$\pm$0.0 & 100.0$\pm$0.0 & 100.0$\pm$0.0 & 100.0$\pm$0.0 & 100.0$\pm$0.0 & 93.7$\pm$2.6 & 57.5$\pm$0.6 & 55.6$\pm$0.9 & 55.2$\pm$0.6 & 64.6$\pm$4.5 \\
\midrule
\multirow{5}{*}{\textbf{GTN}} & Accuracy & 54.6$\pm$0.3 & 56.0$\pm$0.5 & 61.8$\pm$1.2 & 60.8$\pm$1.7 & 59.3$\pm$2.9 & 72.5$\pm$7.9 & 91.5$\pm$2.8 & 94.0$\pm$0.3 & 50.6$\pm$0.6 & 52.7$\pm$1.4 & 73.9$\pm$9.9 & 67.9$\pm$7.7 & 90.6$\pm$0.8 & 91.5$\pm$1.1 & 94.9$\pm$0.9 & 95.4$\pm$0.7 & 94.5$\pm$1.0 & 91.2$\pm$3.5 & 91.6$\pm$3.2 & 92.3$\pm$1.8 & 96.5$\pm$0.7 & 96.5$\pm$0.7 & 96.5$\pm$0.7 & 96.5$\pm$0.7 & 96.5$\pm$0.7 & 96.5$\pm$0.7 & 96.5$\pm$0.7 & 92.6$\pm$1.5 & 54.8$\pm$2.0 & 53.4$\pm$0.6 & 53.4$\pm$0.6 & 58.4$\pm$2.4 \\
  & F1-score & 26.3$\pm$2.2 & 30.1$\pm$0.4 & 44.5$\pm$3.7 & 42.0$\pm$4.5 & 38.3$\pm$7.5 & 64.5$\pm$13.9 & 91.3$\pm$3.2 & 94.1$\pm$0.4 & 14.2$\pm$3.8 & 20.6$\pm$5.6 & 66.1$\pm$16.9 & 55.9$\pm$16.3 & 90.4$\pm$0.8 & 91.4$\pm$1.1 & 95.0$\pm$0.8 & 95.6$\pm$0.7 & 94.6$\pm$1.1 & 90.9$\pm$3.8 & 91.5$\pm$3.4 & 92.2$\pm$1.9 & 96.6$\pm$0.6 & 96.6$\pm$0.6 & 96.6$\pm$0.6 & 96.6$\pm$0.6 & 96.6$\pm$0.6 & 96.6$\pm$0.6 & 96.6$\pm$0.6 & 92.6$\pm$1.7 & 26.7$\pm$4.9 & 23.0$\pm$1.5 & 22.9$\pm$2.5 & 36.4$\pm$4.9 \\
  & Precision & 70.0$\pm$1.8 & 73.1$\pm$3.3 & 81.5$\pm$1.0 & 80.3$\pm$2.2 & 78.1$\pm$4.5 & 87.5$\pm$3.7 & 92.8$\pm$0.8 & 93.2$\pm$1.1 & 53.7$\pm$2.6 & 63.3$\pm$4.3 & 88.1$\pm$2.6 & 85.0$\pm$4.6 & 92.7$\pm$1.3 & 92.8$\pm$1.3 & 93.3$\pm$1.2 & 93.3$\pm$1.2 & 93.2$\pm$1.0 & 92.7$\pm$1.7 & 92.8$\pm$1.6 & 92.9$\pm$1.2 & 93.5$\pm$1.1 & 93.5$\pm$1.1 & 93.5$\pm$1.1 & 93.5$\pm$1.1 & 93.5$\pm$1.1 & 93.5$\pm$1.1 & 93.5$\pm$1.1 & 93.0$\pm$1.0 & 69.9$\pm$6.2 & 66.6$\pm$3.6 & 66.5$\pm$2.7 & 77.1$\pm$5.3 \\
  & Evasion & 83.8$\pm$1.8 & 81.1$\pm$0.4 & 69.3$\pm$3.7 & 71.4$\pm$4.0 & 74.5$\pm$6.1 & 48.0$\pm$15.7 & 10.1$\pm$6.7 & 5.0$\pm$2.0 & \textbf{91.7$\pm$2.5} & \textbf{87.6$\pm$3.9} & 45.3$\pm$21.0 & 57.2$\pm$16.2 & 11.8$\pm$0.4 & 9.9$\pm$1.3 & 3.1$\pm$1.0 & 2.1$\pm$0.8 & 3.9$\pm$3.3 & 10.7$\pm$5.9 & 9.7$\pm$5.3 & 8.4$\pm$3.8 & 0.0$\pm$0.0 & 0.0$\pm$0.0 & 0.0$\pm$0.0 & 0.0$\pm$0.0 & 0.0$\pm$0.0 & 0.0$\pm$0.0 & 0.0$\pm$0.0 & 7.8$\pm$4.2 & 83.5$\pm$3.4 & 86.1$\pm$1.1 & \textbf{86.2$\pm$1.8} & \textbf{76.1$\pm$3.8} \\
  & AUROC & 61.3$\pm$1.1 & 63.4$\pm$1.6 & 71.2$\pm$0.7 & 69.9$\pm$1.9 & 67.7$\pm$2.8 & 83.2$\pm$6.5 & 96.8$\pm$1.9 & 98.7$\pm$0.4 & 51.9$\pm$1.8 & 56.6$\pm$4.0 & 81.8$\pm$8.8 & 76.0$\pm$5.9 & 96.5$\pm$0.3 & 97.0$\pm$0.6 & 99.1$\pm$0.2 & 99.3$\pm$0.3 & 98.8$\pm$1.0 & 96.6$\pm$1.9 & 96.9$\pm$1.8 & 97.1$\pm$1.5 & 100.0$\pm$0.0 & 100.0$\pm$0.0 & 100.0$\pm$0.0 & 100.0$\pm$0.0 & 100.0$\pm$0.0 & 100.0$\pm$0.0 & 100.0$\pm$0.0 & 97.6$\pm$1.4 & 58.0$\pm$1.8 & 56.4$\pm$0.3 & 56.9$\pm$0.2 & 64.2$\pm$3.2 \\
\midrule
\multirow{5}{*}{\textbf{PI-GCN}} & Accuracy & 93.1$\pm$1.7 & 95.8$\pm$1.1 & 96.5$\pm$0.6 & 97.1$\pm$0.6 & 98.2$\pm$0.7 & 98.2$\pm$0.7 & 98.2$\pm$0.7 & 98.2$\pm$0.7 & 80.6$\pm$7.1 & 96.4$\pm$0.6 & 97.0$\pm$0.7 & 95.0$\pm$2.3 & 98.1$\pm$0.7 & 98.2$\pm$0.7 & 98.2$\pm$0.7 & 98.2$\pm$0.7 & 88.5$\pm$13.8 & 92.6$\pm$4.8 & 92.6$\pm$4.9 & 96.3$\pm$2.3 & 98.2$\pm$0.7 & 98.2$\pm$0.7 & 98.2$\pm$0.7 & 98.2$\pm$0.7 & 98.2$\pm$0.7 & 98.2$\pm$0.7 & 98.2$\pm$0.7 & 92.0$\pm$6.5 & 58.3$\pm$4.3 & 54.9$\pm$0.2 & 54.3$\pm$0.5 & 62.2$\pm$3.4 \\
  & F1-score & 92.8$\pm$1.9 & 95.8$\pm$1.2 & 96.5$\pm$0.6 & 97.1$\pm$0.6 & 98.2$\pm$0.7 & 98.2$\pm$0.7 & 98.2$\pm$0.7 & 98.2$\pm$0.7 & 76.3$\pm$10.7 & 96.4$\pm$0.6 & 97.0$\pm$0.7 & 94.9$\pm$2.4 & 98.2$\pm$0.7 & 98.2$\pm$0.7 & 98.2$\pm$0.7 & 98.2$\pm$0.7 & 85.6$\pm$19.0 & 92.2$\pm$5.3 & 92.1$\pm$5.5 & 96.3$\pm$2.4 & 98.2$\pm$0.7 & 98.2$\pm$0.7 & 98.2$\pm$0.7 & 98.2$\pm$0.7 & 98.2$\pm$0.7 & 98.2$\pm$0.7 & 98.2$\pm$0.7 & 91.3$\pm$7.8 & 31.8$\pm$13.0 & 22.9$\pm$1.4 & 21.2$\pm$3.1 & 42.4$\pm$7.3 \\
  & Precision & 96.1$\pm$1.4 & 96.3$\pm$1.4 & 96.4$\pm$1.4 & 96.4$\pm$1.4 & 96.5$\pm$1.4 & 96.5$\pm$1.4 & 96.5$\pm$1.4 & 96.5$\pm$1.4 & 94.7$\pm$1.0 & 96.4$\pm$1.3 & 96.4$\pm$1.3 & 96.2$\pm$1.4 & 96.5$\pm$1.4 & 96.5$\pm$1.4 & 96.5$\pm$1.4 & 96.5$\pm$1.4 & 95.5$\pm$1.5 & 96.0$\pm$1.6 & 96.0$\pm$1.6 & 96.3$\pm$1.3 & 96.5$\pm$1.4 & 96.5$\pm$1.4 & 96.5$\pm$1.4 & 96.5$\pm$1.4 & 96.5$\pm$1.4 & 96.5$\pm$1.4 & 96.5$\pm$1.4 & 96.0$\pm$1.1 & 83.9$\pm$3.7 & 79.0$\pm$5.0 & 77.6$\pm$4.1 & 88.1$\pm$5.4 \\
  & Evasion & 10.2$\pm$3.7 & 4.7$\pm$2.2 & 3.2$\pm$1.2 & 2.2$\pm$0.8 & 0.0$\pm$0.0 & 0.0$\pm$0.0 & 0.0$\pm$0.0 & 0.0$\pm$0.0 & 35.1$\pm$15.5 & 3.4$\pm$2.6 & 2.3$\pm$1.9 & 6.3$\pm$4.5 & 0.0$\pm$0.1 & 0.0$\pm$0.0 & 0.0$\pm$0.0 & 0.0$\pm$0.0 & 19.3$\pm$28.5 & 11.0$\pm$9.4 & 11.2$\pm$9.7 & 3.7$\pm$4.9 & 0.0$\pm$0.0 & 0.0$\pm$0.0 & 0.0$\pm$0.0 & 0.0$\pm$0.0 & 0.0$\pm$0.0 & 0.0$\pm$0.0 & 0.0$\pm$0.0 & 12.2$\pm$14.1 & \textbf{79.8$\pm$10.0} & \textbf{86.6$\pm$1.1} & \textbf{87.6$\pm$2.2} & \textbf{71.9$\pm$6.0} \\
  & AUROC & 95.8$\pm$2.6 & 98.1$\pm$1.1 & 98.6$\pm$0.7 & 99.0$\pm$0.5 & 100.0$\pm$0.0 & 100.0$\pm$0.0 & 100.0$\pm$0.0 & 100.0$\pm$0.0 & 88.0$\pm$8.5 & 99.1$\pm$0.8 & 99.6$\pm$0.4 & 97.8$\pm$2.0 & 100.0$\pm$0.0 & 100.0$\pm$0.0 & 100.0$\pm$0.0 & 100.0$\pm$0.0 & 90.7$\pm$14.7 & 95.5$\pm$4.7 & 95.4$\pm$4.7 & 98.6$\pm$2.1 & 100.0$\pm$0.0 & 100.0$\pm$0.0 & 100.0$\pm$0.0 & 100.0$\pm$0.0 & 100.0$\pm$0.0 & 100.0$\pm$0.0 & 100.0$\pm$0.0 & 98.3$\pm$1.5 & 60.5$\pm$5.7 & 56.6$\pm$3.1 & 55.9$\pm$3.0 & 64.8$\pm$0.6 \\
\midrule
\multirow{5}{*}{\textbf{PI-GDN}} & Accuracy & 93.7$\pm$0.6 & 95.6$\pm$0.6 & 96.0$\pm$0.4 & 96.4$\pm$0.5 & 97.3$\pm$0.5 & 97.3$\pm$0.5 & 97.3$\pm$0.5 & 97.3$\pm$0.5 & 89.8$\pm$1.2 & 96.6$\pm$0.9 & 96.7$\pm$0.9 & 95.4$\pm$1.1 & 97.3$\pm$0.5 & 97.3$\pm$0.5 & 97.3$\pm$0.5 & 97.3$\pm$0.5 & 93.3$\pm$5.0 & 94.6$\pm$2.2 & 94.4$\pm$2.2 & 96.9$\pm$0.5 & 97.3$\pm$0.5 & 97.3$\pm$0.5 & 97.3$\pm$0.5 & 97.3$\pm$0.5 & 97.3$\pm$0.5 & 97.3$\pm$0.5 & 97.3$\pm$0.5 & 97.0$\pm$0.7 & 62.4$\pm$4.1 & 57.1$\pm$3.3 & 55.6$\pm$1.5 & 64.6$\pm$6.2 \\
  & F1-score & 93.7$\pm$0.6 & 95.6$\pm$0.6 & 96.1$\pm$0.3 & 96.4$\pm$0.4 & 97.4$\pm$0.5 & 97.4$\pm$0.5 & 97.4$\pm$0.5 & 97.4$\pm$0.5 & 89.3$\pm$1.2 & 96.7$\pm$0.9 & 96.8$\pm$0.9 & 95.5$\pm$1.1 & 97.4$\pm$0.5 & 97.4$\pm$0.5 & 97.4$\pm$0.5 & 97.4$\pm$0.5 & 93.0$\pm$5.7 & 94.6$\pm$2.3 & 94.4$\pm$2.4 & 96.9$\pm$0.4 & 97.4$\pm$0.5 & 97.4$\pm$0.5 & 97.4$\pm$0.5 & 97.4$\pm$0.5 & 97.4$\pm$0.5 & 97.4$\pm$0.5 & 97.4$\pm$0.5 & 97.1$\pm$0.6 & 44.2$\pm$9.3 & 31.2$\pm$7.6 & 27.3$\pm$3.1 & 48.9$\pm$12.1 \\
  & Precision & 94.6$\pm$0.9 & 94.7$\pm$0.9 & 94.8$\pm$0.9 & 94.8$\pm$0.9 & 94.9$\pm$0.9 & 94.9$\pm$0.9 & 94.9$\pm$0.9 & 94.9$\pm$0.9 & 94.1$\pm$1.1 & 94.9$\pm$0.9 & 94.9$\pm$0.9 & 94.7$\pm$1.0 & 94.9$\pm$0.9 & 94.9$\pm$0.9 & 94.9$\pm$0.9 & 94.9$\pm$0.9 & 94.5$\pm$0.4 & 94.6$\pm$1.0 & 94.6$\pm$0.9 & 94.9$\pm$0.9 & 94.9$\pm$0.9 & 94.9$\pm$0.9 & 94.9$\pm$0.9 & 94.9$\pm$0.9 & 94.9$\pm$0.9 & 94.9$\pm$0.9 & 94.9$\pm$0.9 & 94.9$\pm$0.9 & 84.3$\pm$4.7 & 77.7$\pm$7.4 & 75.5$\pm$5.4 & 85.7$\pm$5.6 \\
  & Evasion & 7.2$\pm$1.1 & 3.5$\pm$0.6 & 2.6$\pm$0.5 & 1.9$\pm$0.2 & 0.0$\pm$0.0 & 0.0$\pm$0.0 & 0.0$\pm$0.0 & 0.0$\pm$0.0 & 15.0$\pm$1.5 & 1.4$\pm$1.6 & 1.2$\pm$1.3 & 3.8$\pm$1.8 & 0.0$\pm$0.0 & 0.0$\pm$0.0 & 0.0$\pm$0.0 & 0.0$\pm$0.0 & 8.1$\pm$11.0 & 5.4$\pm$4.4 & 5.8$\pm$4.6 & 0.9$\pm$0.7 & 0.0$\pm$0.0 & 0.0$\pm$0.0 & 0.0$\pm$0.0 & 0.0$\pm$0.0 & 0.0$\pm$0.0 & 0.0$\pm$0.0 & 0.0$\pm$0.0 & 0.7$\pm$0.5 & \textbf{69.8$\pm$7.8} & \textbf{80.4$\pm$5.6} & \textbf{83.3$\pm$2.1} & \textbf{65.4$\pm$11.4} \\
  & AUROC & 97.4$\pm$0.4 & 98.6$\pm$0.2 & 98.9$\pm$0.1 & 99.2$\pm$0.1 & 100.0$\pm$0.0 & 100.0$\pm$0.0 & 100.0$\pm$0.0 & 100.0$\pm$0.0 & 95.9$\pm$0.6 & 99.7$\pm$0.4 & 99.7$\pm$0.3 & 98.9$\pm$0.5 & 100.0$\pm$0.0 & 100.0$\pm$0.0 & 100.0$\pm$0.0 & 100.0$\pm$0.0 & 97.1$\pm$4.0 & 98.2$\pm$1.5 & 98.1$\pm$1.6 & 99.7$\pm$0.2 & 100.0$\pm$0.0 & 100.0$\pm$0.0 & 100.0$\pm$0.0 & 100.0$\pm$0.0 & 100.0$\pm$0.0 & 100.0$\pm$0.0 & 100.0$\pm$0.0 & 99.8$\pm$0.2 & 67.5$\pm$4.3 & 62.2$\pm$2.7 & 59.4$\pm$1.1 & 70.5$\pm$6.5 \\
\midrule
\multirow{5}{*}{\textbf{PI-GTN}} & Accuracy & 94.1$\pm$0.7 & 96.0$\pm$0.8 & 96.2$\pm$0.8 & 96.5$\pm$1.0 & 97.2$\pm$1.1 & 97.2$\pm$1.1 & 97.2$\pm$1.1 & 97.2$\pm$1.1 & 90.9$\pm$1.6 & 96.6$\pm$0.5 & 96.8$\pm$0.8 & 95.5$\pm$1.3 & 97.2$\pm$1.1 & 97.2$\pm$1.1 & 97.2$\pm$1.1 & 97.2$\pm$1.1 & 92.2$\pm$5.5 & 94.8$\pm$2.6 & 94.8$\pm$2.6 & 96.6$\pm$1.2 & 97.2$\pm$1.1 & 97.2$\pm$1.1 & 97.2$\pm$1.1 & 97.2$\pm$1.1 & 97.2$\pm$1.1 & 97.2$\pm$1.1 & 97.2$\pm$1.1 & 97.0$\pm$0.9 & 62.9$\pm$5.6 & 57.4$\pm$2.8 & 55.8$\pm$1.5 & 65.0$\pm$6.7 \\
  & F1-score & 94.1$\pm$0.6 & 96.0$\pm$0.7 & 96.3$\pm$0.7 & 96.5$\pm$1.0 & 97.2$\pm$1.0 & 97.2$\pm$1.0 & 97.2$\pm$1.0 & 97.2$\pm$1.0 & 90.5$\pm$2.0 & 96.7$\pm$0.4 & 96.8$\pm$0.7 & 95.5$\pm$1.3 & 97.2$\pm$1.0 & 97.2$\pm$1.0 & 97.2$\pm$1.0 & 97.2$\pm$1.0 & 91.8$\pm$6.3 & 94.8$\pm$2.6 & 94.8$\pm$2.6 & 96.7$\pm$1.2 & 97.2$\pm$1.0 & 97.2$\pm$1.0 & 97.2$\pm$1.0 & 97.2$\pm$1.0 & 97.2$\pm$1.0 & 97.2$\pm$1.0 & 97.2$\pm$1.0 & 97.1$\pm$0.9 & 44.8$\pm$14.8 & 32.2$\pm$7.8 & 27.9$\pm$5.7 & 49.7$\pm$13.3 \\
  & Precision & 94.3$\pm$2.0 & 94.5$\pm$1.9 & 94.6$\pm$1.9 & 94.6$\pm$1.9 & 94.7$\pm$1.9 & 94.7$\pm$1.9 & 94.7$\pm$1.9 & 94.7$\pm$1.9 & 94.0$\pm$1.9 & 94.6$\pm$1.9 & 94.6$\pm$1.9 & 94.5$\pm$2.0 & 94.7$\pm$1.9 & 94.7$\pm$1.9 & 94.7$\pm$1.9 & 94.7$\pm$1.9 & 94.1$\pm$1.8 & 94.4$\pm$2.1 & 94.4$\pm$2.1 & 94.6$\pm$1.9 & 94.7$\pm$1.9 & 94.7$\pm$1.9 & 94.7$\pm$1.9 & 94.7$\pm$1.9 & 94.7$\pm$1.9 & 94.7$\pm$1.9 & 94.7$\pm$1.9 & 94.6$\pm$1.9 & 84.2$\pm$3.3 & 78.2$\pm$6.0 & 75.5$\pm$4.3 & 85.6$\pm$6.5 \\
  & Evasion & 6.1$\pm$1.4 & 2.4$\pm$0.5 & 1.9$\pm$0.6 & 1.4$\pm$0.2 & 0.0$\pm$0.0 & 0.0$\pm$0.0 & 0.0$\pm$0.0 & 0.0$\pm$0.0 & 12.6$\pm$5.0 & 1.1$\pm$1.2 & 0.8$\pm$0.8 & 3.4$\pm$1.0 & 0.0$\pm$0.0 & 0.0$\pm$0.0 & 0.0$\pm$0.0 & 0.0$\pm$0.0 & 9.9$\pm$11.7 & 4.7$\pm$3.6 & 4.7$\pm$3.8 & 1.2$\pm$0.7 & 0.0$\pm$0.0 & 0.0$\pm$0.0 & 0.0$\pm$0.0 & 0.0$\pm$0.0 & 0.0$\pm$0.0 & 0.0$\pm$0.0 & 0.0$\pm$0.0 & 0.3$\pm$0.3 & \textbf{68.6$\pm$12.8} & \textbf{79.5$\pm$5.9} & \textbf{82.8$\pm$4.2} & \textbf{64.3$\pm$13.0} \\
  & AUROC & 98.1$\pm$0.2 & 99.2$\pm$0.1 & 99.4$\pm$0.2 & 99.5$\pm$0.2 & 100.0$\pm$0.0 & 100.0$\pm$0.0 & 100.0$\pm$0.0 & 100.0$\pm$0.0 & 96.9$\pm$0.3 & 99.8$\pm$0.2 & 99.8$\pm$0.2 & 99.0$\pm$0.5 & 100.0$\pm$0.0 & 100.0$\pm$0.0 & 100.0$\pm$0.0 & 100.0$\pm$0.0 & 97.0$\pm$4.0 & 98.4$\pm$1.3 & 98.3$\pm$1.3 & 99.7$\pm$0.2 & 100.0$\pm$0.0 & 100.0$\pm$0.0 & 100.0$\pm$0.0 & 100.0$\pm$0.0 & 100.0$\pm$0.0 & 100.0$\pm$0.0 & 100.0$\pm$0.0 & 99.8$\pm$0.2 & 69.8$\pm$4.9 & 64.1$\pm$2.9 & 61.4$\pm$1.1 & 72.3$\pm$6.9 \\
\midrule
\textbf{BDD $\chi^2$} & Evasion & 50.3$\pm$1.0 & 33.5$\pm$1.1 & 28.1$\pm$0.8 & 25.2$\pm$0.5 & 2.5$\pm$1.5 & 0.0$\pm$0.0 & 0.0$\pm$0.0 & 0.0$\pm$0.0 & 74.7$\pm$3.2 & 30.0$\pm$5.7 & 28.3$\pm$8.1 & 43.0$\pm$8.6 & 0.3$\pm$0.1 & 0.0$\pm$0.0 & 0.0$\pm$0.0 & 0.0$\pm$0.0 & 42.6$\pm$35.2 & 47.6$\pm$17.3 & 47.8$\pm$16.7 & 36.9$\pm$12.5 & 0.0$\pm$0.0 & 0.0$\pm$0.0 & 0.0$\pm$0.0 & 0.0$\pm$0.0 & 0.0$\pm$0.0 & 0.0$\pm$0.0 & 0.0$\pm$0.0 & \textbf{95.8$\pm$2.7} & \textbf{82.4$\pm$4.7} & \textbf{86.8$\pm$2.5} & \textbf{87.9$\pm$1.0} & 76.6$\pm$8.7 \\
\midrule
\textbf{BDD LNR} & Evasion & 57.4$\pm$1.0 & 48.0$\pm$0.8 & 46.0$\pm$0.2 & 44.0$\pm$0.2 & 1.4$\pm$0.7 & 0.1$\pm$0.1 & 0.0$\pm$0.0 & 0.0$\pm$0.0 & 54.2$\pm$4.6 & 20.2$\pm$10.0 & 23.2$\pm$13.2 & 38.0$\pm$8.0 & 0.0$\pm$0.0 & 0.0$\pm$0.0 & 0.0$\pm$0.0 & 0.0$\pm$0.0 & 27.4$\pm$27.6 & 28.1$\pm$16.4 & 28.3$\pm$16.0 & 18.7$\pm$8.8 & 0.0$\pm$0.0 & 0.0$\pm$0.0 & 0.0$\pm$0.0 & 0.0$\pm$0.0 & 0.0$\pm$0.0 & 0.0$\pm$0.0 & 0.0$\pm$0.0 & 42.7$\pm$14.6 & \textbf{81.2$\pm$5.9} & \textbf{89.7$\pm$3.0} & \textbf{91.3$\pm$1.7} & \textbf{81.9$\pm$7.0} \\
\bottomrule
\end{tabular}%
}
\end{table*}

Contiguous observed-hour blocks were assigned to training (70\%), validation (10\%), and testing (20\%), with each block reserving 24 hours of input history to prevent overlap between sets. The evaluated test set contained 1,149--1,152 hours per seed. Models and normalization used training data only, the validation split determined the detection thresholds, and seeds 41, 42, and 43 provided independent trials. Each detector was trained with balanced benign and attack examples from the eight families; baseline attack generators retained their respective training budgets. Attacks were frozen before evaluation on matched held-out hours. Contiguous 72-hour blocks were assigned reproducibly by seed, with all detection thresholds fixed before testing. The evaluation was offline, without attacker access to topology, actual bus-demand trajectories, or detector outcomes.

\emph{Detection measures.} Accuracy, precision, F1-score, area under the receiver operating characteristic curve (AUROC), and evasion ($1-\mathrm{recall}$) are computed at validation-calibrated 5\%-FPR operating points, except AUROC, which is threshold independent. Test FPR is measured rather than assumed to equal 5\%. Chi-square and LNR BDD use the same AC-WLS measurements and matched test hours; their convergence and invalid-score counts are monitored separately. In addition to the nominal 95\% chi-square cutoff, the data-driven comparison uses the 95th percentile of benign validation WLS statistics; LNR likewise uses a benign-validation cutoff targeting 5\% FPR. These decisions are computed without attack-specific threshold tuning. Reported BDD evasion is conditioned on valid converged state estimates; an unavailable WLS solution is not automatically counted as residual-test evasion.

\label{sec:spatial_load}
\emph{Physical consequences.}
We assess the physical impact of FDI attacks primarily through spatial load misperception (SLM), which quantifies distortions in the estimated distribution of bus-level demand. Such distortions can affect dispatch, congestion monitoring, and reserve allocation while remaining difficult to detect through aggregate-demand checks when errors across buses cancel. For matched converged benign and attacked AC-WLS estimates, let $\boldsymbol\delta_t=\widehat{\mathbf P}^{\mathrm{d,a}}_t-\widehat{\mathbf P}^{\mathrm{d,b}}_t$ denote the difference in estimated demand vectors. SLM is defined as
\begin{equation}
\label{eq:spatial_impact}
\Lambda_t=\|\boldsymbol\delta_t\|_1,\qquad
\Lambda_t^{\%}=100\Lambda_t/\textstyle\sum_{i\in\mathcal B}P_{i,t}^{\mathrm d},
\end{equation}
where $P_{i,t}^{\mathrm d}$ denotes the actual bus demand available only to the evaluator. We also measure potential next-ED bias arising from changes in perceived total demand. With $\mathbf1$ denoting the all-ones vector and $\Delta P_t=\mathbf1^\top\boldsymbol\delta_t$, the corresponding overgeneration and undergeneration indicators are $\max(\Delta P_t,0)$ and $\max(-\Delta P_t,0)$, respectively, with zero-bias hours included. The  stealth-impact trade-off score multiplies mean $\Lambda^{\%}$ by detector evasion for each seed, scenario, and attack before averaging across seeds. It does not represent an hour-wise joint success probability. AC-WLS nonconvergence is reported as disruption rather than residual-test evasion.

\begin{figure*}[!t]
    \centering
    \subfloat[\label{fig:spatial_impact}]{%
        \includegraphics[width=0.5\textwidth]{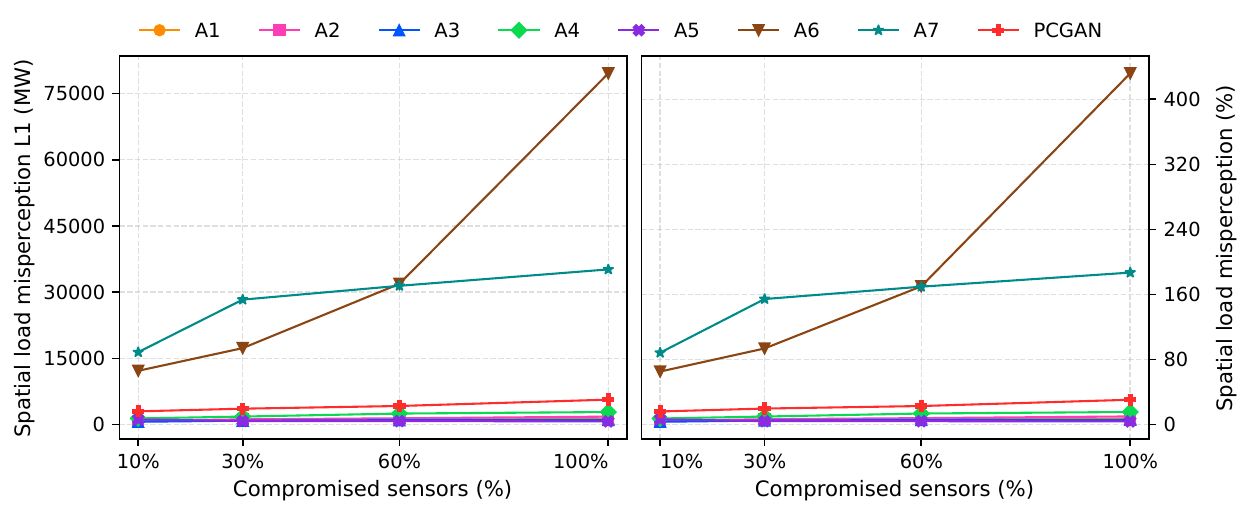}%
    }%
    \subfloat[\label{fig:dispatch_bias}]{%
        \includegraphics[width=0.5\textwidth]{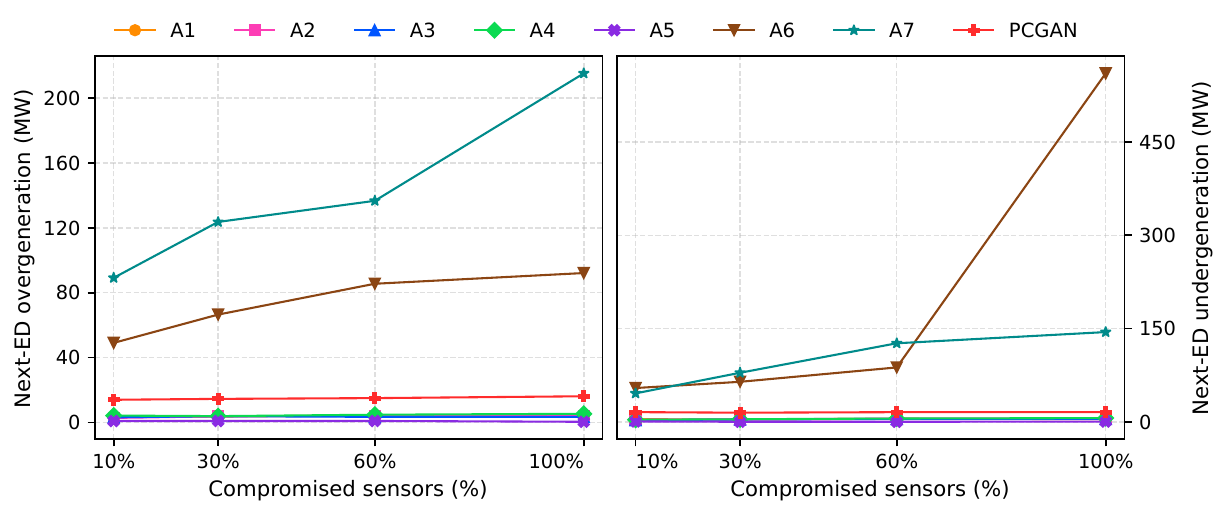}%
    }
    \caption{Physical consequences of FDI attacks.
    (a) Spatial load misperception;
    (b)  demand-generation mismatch occurs in subsequent dispatch.}
    \label{fig:physical_consequences}
\end{figure*}

\begin{figure*}[!t]
\centering
\includegraphics[width=0.99\textwidth]{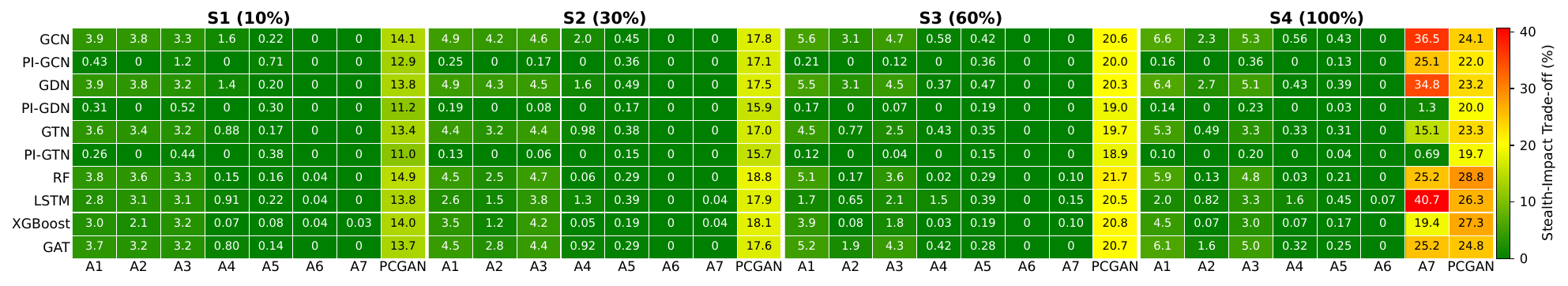}
\caption{Stealth--impact trade-off score in terms of SLM $\times$ ML Evasion (\%) in S1--S4; A1--A7 are benchmark attacks.}
\label{fig:tradeoff_map}
\end{figure*}

\section{Results and Discussion}
\label{sec:results_discussion}

We evaluate the effectiveness of PCGAN against competing FDI attacks in terms of detection evasion and physical impact under varying sensor-compromise levels.

\emph{Attack Stealth.}
Table~\ref{tab:detector_all_attacks_pjm_5pct} presents five metrics for eight attacks and ten ML detectors across S1--S4, along with evasion rates for two AC-WLS BDD methods. Entries are reported as mean $\pm$ SD across three seeds, using thresholds calibrated to a 5\% validation FPR. All results are obtained from the implemented benchmarks under a common evaluation protocol.

At the common validation-calibrated operating points, PCGAN attains 84.28\% mean ML evasion, exceeding BIM, the next-highest competitor, at 57.08\%. Its detector-averaged evasion is 82.35\%, 87.88\%, 88.56\%, and 78.31\% in S1--S4, respectively. Among the detectors, PI-GTN provides the strongest overall detection performance, with only 10.80\% mean evasion across all attacks and scenarios, followed by PI-GDN (11.14\%) and PI-GCN (14.09\%). Notably, BIM and RA achieve 90.02\% and 89.75\% evasion against GCN, respectively, but only 11.78\% and 5.07\% against PI-GCN. Similar improvements are observed for the other physics-informed graph models, indicating that physical consistency features help reveal attacks that conventional graph-based detectors frequently miss. Nevertheless, PCGAN maintains 73.80\% evasion even against PI-GTN. Averaging, over scenarios and seeds, the lowest evasion recorded against any single detector in each run, PCGAN retains 72.70\%, compared with 4.22\% for BIM, demonstrating its effectiveness against heterogeneous detection architectures. PCGAN's mean AUROC is 59.39\%, while its observed test FPR averages 5.36\% at the nominal 5\%-validation-FPR thresholds.

The classical tests show a similar pattern. PCGAN achieves 83.45\% chi-square BDD evasion and 86.02\% LNR BDD evasion. Chi-square evasion remains 82.45\%, 86.79\%, and 87.89\% in S1--S3 despite protected measurement channels. This behavior is consistent with low residual visibility, although it does not verify the sufficient conditions of Theorem~\ref{thm:random}. In contrast, RL-iGAN reaches 95.80\% chi-square evasion at S4 but only 11.71\% ML evasion; its chi-square evasion is zero under partial access.  These differences show why residual BDD and learned detectors must be evaluated separately. BDD checks whether the received measurements are electrically consistent rather than how far the estimated state is from the true state. Therefore, ML detectors can still flag the temporal and statistical anomalies that BDD misses. Their rankings may consequently disagree even when both use identical attacked samples. FDICGAN is  strongly detected by both detector families despite generating large estimated-load deviations. PCGAN's approximate power-flow guidance offers a plausible explanation for its residual-test performance, while its adversarial training promotes similarity to benign histories.

\emph{Attack Impact.}
Fig.~\ref{fig:physical_consequences}(a) illustrates SLM across the four scenarios. As attacker access increases, the set of writable measurements expands. Accordingly, PCGAN's SLM rises from 16.14\% (S1) to 19.73\% (S2), 22.83\% (S3), and 30.55\% (S4), averaging 22.31\% (approximately 4.11~GW), with an across-seed SD of 0.17 percentage points. However, ML evasion peaks at S3 before declining in S4, indicating that greater impact does not necessarily imply greater stealth. LRA produces 11.63\% mean SLM. FDICGAN and RL-iGAN produce substantially larger estimated distortions (190.20\% and 149.83\%), but their low ML evasion limits their effectiveness as stealthy attacks. AC-WLS convergence does not guarantee the physical plausibility of these estimates.

Fig.~\ref{fig:physical_consequences}(b) shows PCGAN's mean potential next-dispatch overgeneration and undergeneration of 14.89 and 15.47~MW, respectively. These values are considerably smaller than its absolute SLM because positive and negative bus-level estimation errors partially cancel when aggregated. Thus, the attack can substantially distort the perceived spatial demand distribution while producing a relatively small aggregate demand bias. These indicators represent potential dispatch errors rather than realized generation changes, as downstream dispatch was not simulated. All recorded benign and attacked AC-WLS evaluations converged, with no attack-induced estimation failures.

\emph{Stealth--impact trade-off.} Fig.~\ref{fig:tradeoff_map} combines the two preceding dimensions through the seed-matched product of SLM and ML evasion. PCGAN's detector-averaged score increases from 13.28\% in S1 to 23.93\% in S4 and averages 18.69\% overall. RL-iGAN has the highest competing overall score at 5.61\%, largely because its S4 score reaches 22.40\%; under S1--S3, its score is negligible. The narrow S4 gap therefore does not extend to restricted-access scenarios. RA, BIM, GA, and LRA attain overall scores of 3.12\%, 2.74\%, 1.52\%, and 0.48\%, respectively. Their lower combined scores reflect insufficient evasion, limited SLM, or both. PCGAN offers the strongest average trade-off across the four sensor-access levels. The score is descriptive and does not represent the probability of simultaneous stealth and impact.

\section{Conclusion}
\label{sec:conclusion}

Existing FDI attack generators assume sensors at every bus and often write access to all of them, require the grid topology, line parameters, or Jacobian, depend on the parameters or gradients of the targeted detector, and report stealth or estimation error rather than physical consequences. This paper addressed these four limitations with a detector-agnostic PCGAN. Sensors are deployed only at selected critical buses, and the attacker writes only to the channels of the compromised sensors. The attacker identifies approximate decoupled power-flow sensitivities from the historical measurements of those sensors alone, without the admittance matrix, the estimator, or the detector parameters, and shapes the written perturbation with a consistency penalty and adversarial training. Stealth and physical impact are assessed jointly through ten heterogeneous ML detectors, residual-based BDD tests, spatial load misperception under nonlinear AC-WLS estimation, and a stealth--impact trade-off score that combines the two.

On the IEEE 300-bus system with 10--100\% of the deployed sensors compromised, PCGAN averages 84.28\% ML evasion and 22.31\% spatial load misperception, retains 73.80\% evasion against the strongest physics-informed detector and above 83\% BDD evasion, and attains a trade-off score of 18.69\% compared with 5.61\% for the strongest competitor, whereas the benchmark attacks are either stealthy with little impact or impactful but easily detected. The evaluation is offline on a single test system with one year of demand. Future work will test PCGAN on multiple test systems and unseen demand patterns and will develop stronger detectors against PCGAN. This study can help researchers and grid operators assess detector vulnerabilities and improve the robustness of future detection methods against such attacks.

\bibliographystyle{IEEEtran}
\begingroup\renewcommand{\baselinestretch}{0.94}\bibliography{references}\endgroup

\end{document}